# Catastrophic Forgetting in Deep Learning: A Comprehensive Taxonomy

**Everton L. Aleixo**  EVERTON.ALEIXO@ICOMP.UFAM.EDU.BR
**Juan G. Colonna**  JUANCOLONNA@ICOMP.UFAM.EDU.BR
**Marco Cristo**  MARCO.CRISTO@ICOMP.UFAM.EDU.BR
*Universidade Federal do Amazonas, Institute of Computation, Manaus, Amazonas, Brazil*

**Everlandio Fernandes**  EVERLANDIO.FERNANDES@SIDIA.COM
*Sidia Institute of Science and Technology, Manaus, Amazonas, Brazil*

## Abstract

Deep Learning models have achieved remarkable performance in tasks such as image classification or generation, often surpassing human accuracy. However, they can struggle to learn new tasks and update their knowledge without access to previous data, leading to a significant loss of ac- curacy known as Catastrophic Forgetting (CF). This phenomenon was first observed by McCloskey and Cohen in 1989 and remains an active research topic. Incremental learning without forgetting is widely recognized as a crucial aspect in building better AI systems, as it allows models to adapt to new tasks without losing the ability to perform previously learned ones. This article surveys recent studies that tackle CF in modern Deep Learning models that use gradient descent as their learn- ing algorithm. Although several solutions have been proposed, a definitive solution or consensus on assessing CF is yet to be established. The article provides a comprehensive review of recent solutions, proposes a taxonomy to organize them, and identifies research gaps in this area.

## 1. Introduction

Incremental Learning, also known as Continual Learning, is the ability to progressively learn new tasks, one at a time, without forgetting the previously learned tasks, and also use the accumulated knowledge to facilitate learning future tasks. For humans, the process of learning a new task can be simpler when it is related to an already-known task. For instance, learning to ride a motorcycle should be simpler if you already know how to ride a bicycle. This property is called Forward Knowledge Transfer (Mai et al., 2021a), and this is a desirable property for any AI system. Forward Knowledge is widely known as transfer learning with fine-tuning in Neural Networks (Ka¨ding et al., 2016; Zhuang et al., 2021). However, the problem with transfer learning approaches is that prior knowledge is forgotten when learning a new task, causing a drastic reduction in performance on previous tasks. Backward knowledge propagation is another important human characteristic that is desirable in an AI incremental learning system (Ke et al., 2020). This property expects that after learning a task related to a previous one, the performance of the previous one will also increase or at least be maintained. When these properties fail and there is a high decrease in performance on the previous tasks in favor of the new one, the model is said to suffer from Catastrophic Forgetting (CF).

McCloskey and Cohen (1989) showed that the biggest challenge for AI systems to support Incremental Learning is the CF phenomenon. It was studied in many types of models, for instance, Support Vector Machine (SVM) (Ayad, 2014), however, this issue is more notable in connectionist models (French, 1999), and this is mainly caused by the way the parameters of these models are

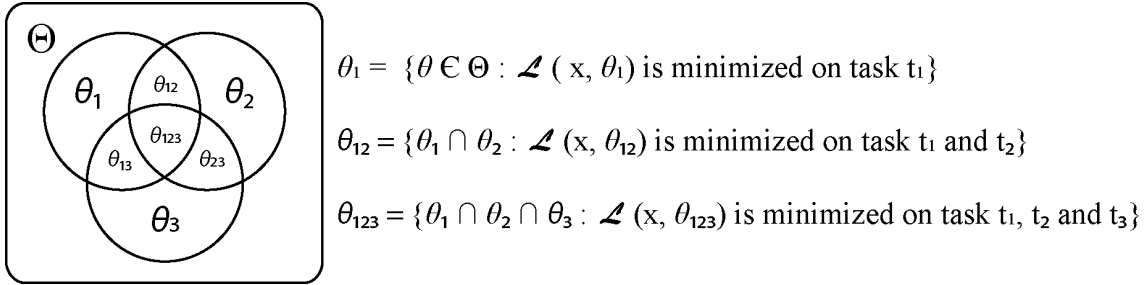

$\theta_1 = \{\theta \in \Theta : \mathcal{L}(x, \theta_1)$ is minimized on task $t_1\}$

$\theta_{12} = \{\theta_1 \cap \theta_2 : \mathcal{L}(x, \theta_{12})$ is minimized on task $t_1$ and $t_2\}$

$\theta_{123} = \{\theta_1 \cap \theta_2 \cap \theta_3 : \mathcal{L}(x, \theta_{123})$ is minimized on task $t_1$, $t_2$ and $t_3\}$

Figure 1: Combinations of Parameter Sets for Simultaneous Task Solving. A subset $\theta_t$ is considered valid if it satisfies a minimum loss criteria denoted by $L(x_i, \theta_t)$.

adjusted. When a new pattern is presented to the model, these parameters — also known as weights — change in value, losing the model's capability to detect previous patterns. There is still no agreement on how to handle this issue in AI today, making it an open problem.

Since the mid-2000s, connectionist models known as Deep Learning (DL) (Goodfellow et al., 2016), have become dominant in Machine Learning. The most common DL models are those constructed from multilayer Artificial Neural Networks (ANNs), Convolution Neural Networks (CNNs), Recurrent Neural Networks (RNNs), and Transformers, which we also refer to as Deep Neural Networks (DNN). Although several DNN models have been proposed, including models that can outperform humans in numerous tasks, such models continue to suffer from the effects of CF on Incremental Learning. The Incremental Learning problem is formulated in several ways, some accepting more relaxed constraints and others trying to better approximate real-world constraints (Lee et al., 2017a; Caccia et al., 2020).

In a recent study by Knoblauch et al. (2020), it was demonstrated that the problem of determining the optimal parameters to avoid the CF in a fixed model can be reduced to the well-known Satisfiability (SAT) problem (Schaefer, 1978). Consequently, this proof categorizes the CF problem within the class of NP-HARD problems. Essentially, each potential arrangement of weights, de- noted as $\theta_t$, within a model can be seen as an element in a set of solutions that satisfy a given task $t$. Thus, the key to solving the CF problem lies in identifying a specific combination of $\theta$ values that exist at the intersection of the solution sets for all tasks. By reducing CF to an NP-HARD problem, it becomes evident that effectively addressing this challenge is crucial for advancements in artificial intelligence in the years to come. In Figure 1, we can visualize the set of all possible parameter combinations, denoted as $\Theta$, for a given model. The subsets $\theta_1$, $\theta_2$, and $\theta_3$ represent the parameter combinations that individually solve tasks 1, 2, and 3, respectively. Furthermore, there are smaller subsets, such as $\theta_{12}$, $\theta_{23}$, and $\theta_{13}$, that solve combinations of two tasks simultaneously. Finally, the set $\theta_{123}$ represents the parameter combination that simultaneously solves all three tasks. It is worth noting that finding this specific combination becomes increasingly challenging as it involves satisfying the requirements of multiple tasks simultaneously, leading to a combinatorial problem.

Despite the NP-HARD nature of CF, significant progress has been made in mitigating CF within Deep Neural Network (DNN) models through various techniques and heuristics. Several studies have proposed effective solutions, including Rusu et al. (2016), Rebuffi et al. (2017), Zenke et al. (2017), Kirkpatrick et al. (2017), Fernando et al. (2017), Roy et al. (2020), Prabhu et al. (2020), Cha et al. (2021), Wang et al. (2022a), Smith et al. (2023). In this paper, we present a comprehensive examination of these techniques, shedding light on the current state of research in this domain.

Due to the absence of a clear consensus on how to evaluate these models numerically, we refrain from direct numerical comparisons. Instead, our focus lies in discussing the most suitable application scenarios for each technique and comprehensively assessing their respective benefits and drawbacks. This approach allows for a more nuanced understanding of the strengths and limitations of each method, aiding researchers and practitioners in making informed decisions regarding their implementation in real-world settings. The primary contributions of this research endeavor can be concisely encapsulated in the following threefold:

- proposes a taxonomy to cluster techniques to avoid CF in DNN models based on the strategy;

- explores the progression of techniques within each category, starting from 2012 with the rise in popularity of DNNs;

- and, discuss the strengths and weaknesses inherent in each category, fostering a comprehensive understanding of their respective merits and limitations.

The rest of this work is organized as follows: in Section 2 of this article, we present related articles that empirically contrast various strategies for preventing catastrophic forgetting (Parisi et al., 2019; Pfülb & Gepperth, 2019; Belouadah et al., 2021; Masana et al., 2020); in Section 3 we pro- pose a taxonomy for DNN models categorized into four main groups: Rehearsal, Distance-Based, Dynamic Networks, and Sub-Networks, according to their learning strategies. These models can be used in many contexts, such as computer vision, natural language processing, and others; and, finally, Section 4 brings final considerations.

## 2. Related work

Parisi et al. (2019) carry out a review on aspects of Incremental Learning with Neural Networks, which also details techniques related to Incremental Learning. Although this work was published recently, several new techniques, insights, and solutions have emerged since then. Therefore, we use a systematic review approach to fully map the latest emerging techniques. We follow the guidelines given by Kitchenham and Charters (2007) to retrieve and filter works considered state-of-the-art in this area. We intend to present and explain the methods published since 2012, as it is the period of greatest attention for DNN models.

The current state of research on CF is limited in that many surveys only compare empirical findings or are confined to a single experimental setup. For example, recent surveys by Masana et al. (2020) and Mai et al. (2021a) have only focused on specific aspects of continual learning. However, De Lange et al. (2021) conducted an extensive study on task incremental learning setups for image classification, where the model is aware of the task it is evaluating. They proposed a taxonomy that categorizes solutions into three categories: Replay, Regularization-based, and Parameter isolation methods, each with two or three sub-categories. Notably, this taxonomy does not include hybrid or distance-based methods, which could be an interesting avenue for future research.

The work by Belouadah et al. (2021) provides a comprehensive analysis of fixed-size models in the context of image classification. Their study focuses specifically on class incremental learning, a setup in which the model has no knowledge of which task an input belongs to when evaluating, but each task introduces new classes. To handle this, models typically create new output neurons for each new class. The authors demonstrate that a straightforward fine-tuning approach using a memory of past tasks, also known as a Replay Buffer, can achieve competitive accuracy compared to more complex methods.

In their study, van de Ven and Tolias (2019) introduce a novel setup called domain-incremental learning, which they investigate in the context of computer vision. In this setup, each new task involves inputs from new classes related to previous ones. For example, the first task may involve detecting cats in images of cats in a garden in the morning, while the second task involves detecting cats in images of cats indoors. The problem remains the same, but the domain changes.

We not only cover computer vision but also a diverse range of topics, such as Natural Language Processing (NLP) and Reinforcement Learning (RL), employing a variety of models including fully connected networks, Generative Neural Networks (GAN), and others. Our survey also presents three setups that have gained attention in online and few-shot learning: (i) Online Incremental Learning,
(ii) Unbounded Task Incremental Learning, and (iii) Data-free Incremental Learning, which have stricter constraints and require methods that are specific to these setups. Previous surveys have not covered these setups, and as demonstrated by Mai et al. (2021a), methods not designed for online learning are inadequate for these setups.

The survey of Masana et al. (2020) shows the four main reasons why CF occurs: i) weight drift, which occurs when the model adjusts its parameters to new tasks; ii) activation drift, similar to the former, but just in the classification layer; iii) inter-task confusion, which occurs when samples from different tasks have similar features; and finally, iv) task-recency bias, which occurs when the model is biased toward the last task trained on. It is important to emphasize that these reasons are valid for the setup of class incremental learning that allows the storage of a part of the data set from previous tasks.

In the context of Natural Language Processing (NLP), there are also efforts to train models incrementally. Biesialska et al. (2020) examined and compared methods used in various NLP appli- cations. In turn, Wu et al. (2022) compares the resilience of pretrained language models to avoid CF in a sequence of tasks using several techniques. However, unlike us, they do not fit current methods into a taxonomy. Instead, they present an evaluation of methods in continuous learning scenarios.

As Kemker et al. (2018) demonstrated, comparing results between models can be influenced by various variables such as task order or chosen setup. Therefore, our goal is not to provide another empirical survey but rather to offer a comprehensive overview of the field from multiple perspectives. To achieve this, we will introduce a taxonomy that categorizes existing methods, allowing for easier comparison between them. In the following section, we will present the most relevant works and organize them according to this taxonomy. Our aim is to provide a broad understanding of the area and understand the appropriate usage of them.

## 3. Taxonomic Organization

To provide a clear framework for organizing methods that prevent CF, we suggest categorizing them into four main groups: rehearsal, distance-based, sub-networks, and dynamic networks. Figure 2 visually represents this taxonomy.

A method falls under the **Rehearsal** category if it utilizes data from previously learned tasks, such as a replay buffer in RL. This can include artificial data or representations of learned knowledge, like embeddings from prior tasks. Models that employ similarity estimates based on the distance between samples to determine task or class membership are classified as **Distance-based**. For instance, samples projected on a hyperplane can be considered part of the same class. The

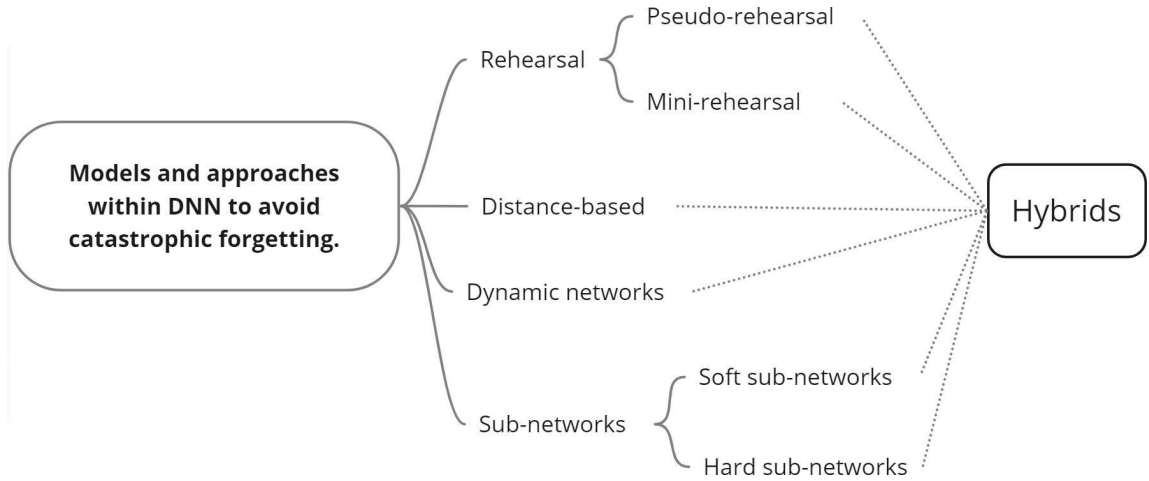

Figure 2: The proposed taxonomy of DNN models in continuous learning. We propose four main categories: i) Rehearsal; ii) Distance based; iii) Sub-networks; and, iv) Dynamic networks. Moreover, for existing models that share properties of more than one group, we classify these ones as Hybrid models. They all try to improve continuous learning when using deep neural networks.

**Sub-networks** group comprises models that aim to prevent knowledge or parameter overlap between tasks. One approach is to divide the model into smaller sub-models, each with its own set of weights, which are used to learn one task. Two sub-models can also share some components. **Dynamic Networks**, on the other hand, refer to models whose structure, such as the number of nodes or layers, expands to accommodate additional tasks. As more tasks are learned, the model's capacity increases, allowing it to recognize more patterns. In the following sections, we will explore each of these categories in greater detail.

### 3.1 Rehearsal

Rehearsal is the process of repeating a previous task while learning a new one to prevent forgetting. This technique is commonly used in human psychology and was introduced in the study of neural networks by Robins Robins (1993, 1995). However, storing all of the data that a model is presented with throughout its lifespan and using it to retrain the model can be impractical, as it requires a large amount of storage space and can significantly increase the amount of time required for training. To address these issues, some researchers have focused on reducing the amount of data that needs to be stored, while others have focused on reducing the training time (Hu et al., 2019; Leontev et al., 2019; Sprechmann et al., 2018; Borsos et al., 2020). These approaches are known as mini-rehearsal and pseudo-rehearsal and will be discussed in the following subsections.

#### 3.1.1 PSEUDO-REHEARSAL

Pseudo-rehearsal is based on the assumption that storing previously seen data is not required to deal with CF. It is sufficient to generate new data synthetically, known as pseudo-samples, in order to represent old classes on demand. As illustrated in Figure 3, the generated data does not have to be identical to the original. In this figure, the first row shows noisy images generated by Mellado et al. (2017) to represent classes from the MNIST (LeCun et al., 1998) data set, namely 0, 1, 2, 3, and 4. Although the samples in the first row cannot be identified as digits by a human, they are intended to activate the same set of neural network connections as the original ones in the second row. Therefore, they can be used to retrain the model from time to time, remembering previously learned concepts.

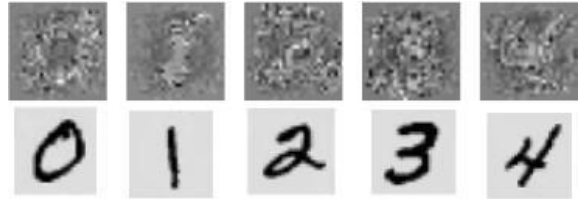

Figure 3: The first row contains noise images that represent classes from 0 to 4 in the neural network trained by Mellado et al. (2017). The second row contains images of classes from 0 to 4 extracted from MNIST.

To perform a new training session and acquire new knowledge, a copy of the existing model is created. This is called a clone of the model. Then, some noisy generated images are labeled by this cloned model. The pairs of noisy images and their labels which have high confidence by the cloned model are concatenated to the current task data set as synthetic data. Thus, the original model is trained on this extended data set. This aims to grant the model to remember classes from an old task. The key problem with synthetically produced samples is that they don't actually belong to the class to which they were assigned. This is known as the ambiguities problem in Incremental Learning (Masana et al., 2020). Therefore, they can conflict with the newest classes the model is learning. Just as a playful example, assume that a neural network was trained to classify images of cats and dogs. Then a new class is added to this neural network. Now assume that the random image generated by the algorithm depicts a sample from the new class (a bird, for example). This image will be labeled as a dog or a cat because the last trained model only knows these two classes. Bayesian Learning and Stochastic Langevin Dynamics were used by Leontev et al. (2019) to
improve the generation of such synthetic data samples. These two techniques were used to generate samples that would maximize the neuron activation, instead of random generation. They forced the generated samples to increase the node output that represents their classes. However, these techniques can lead to the generation of synthetic data with low diversity. To handle this problem the authors proposed a Brownian motion (Karatzas & Shreve, 1991) technique to avoid the selection of similar samples to the data set (using the distance among them in input space). Therefore, they can produce a diverse synthetic data set, discarding the most similar. This technique proved to be resilient to CF in toy data sets, like MNIST, where the dimension of the input data, *(28, 28)* pixels, is not as big as it could be in real applications, such as facial recognition (Schroff et al., 2015). In applications where the input space is larger, the performance of this technique tends to quickly decrease.

The above technique aims to generate synthetic data that will activate specific neurons that

rep- resent certain classes. Nevertheless, there is no assurance that the most elevated ones will be generated. To force this, Smith et al. (2021) propose to use the inverted model trained on the previous task to generate pseudo-samples passing the class label as input. However, the synthetic data generated still does not seem like the original samples. The strategy was used again by PourKeshavarzi et al. (2022), however, different from the previous, they demonstrated that using the Knowledge Distillation (KD) (Hinton et al., 2015) improves the final result. They call this strategy a memory recovery paradigm. And, recently, Liu et al. (2022) used the DeepInversion (Yin et al., 2020) model to generate the pseudo-samples in the Few-shot Incremental Learning setup.

Generative networks were also adopted to create synthetic samples that were more likely to be similar to the original ones. For instance, Mellado et al. (2017) used a Recurrent Neural Network (RNN) called DRAW (Gregor et al., 2015) to generate synthetic data. Different from traditional RNNs, their model has 2 heads: i) a fully connected NN with a *softmax* activation to classify the in- put; and, ii) a decoder to re-create the input. It showed that Generative Adversarial Networks (Good- fellow et al., 2014b) (GANs) could play an important role in helping to mitigate the CF effects for pseudo-rehearsal. In the same year, Shin et al. (2017) propose the Deep Generative Replay framework which uses a GAN as an auxiliary model used at the start of each training session to generate synthetic data from already known classes. The GAN is trained at the end of each training session using the data set of the current task and the self-generated one.

GANs were also used to produce synthetic data to avoid CF in more general scenarios: i) label-conditioned image generation, where the input is the label of class and the output is an image; and, image-conditioned generation, where the input is an image of a class and the output is an image from the same class, but in a different domain. For instance, a cartoon cat image as input can result in a sketch cat image as output (Zhai et al., 2019). Although generative models can generate synthetic data for your own incremental training, these data are not always high quality, problems like blur usually occur. Therefore, Xiang et al. (2019) propose to generate feature maps that are passed to the classifier's internal layers instead of samples in the original input space. The use of feature maps instead of pseudo-samples to do the classifications is an easier task due to the size of the feature map in comparison to the raw image.

Learning using Encoded Experience Replay (CLEER) (Rostami et al., 2019) is an Autoencoder (AE) trained to make all tasks share the same distribution in the embedding space. For this, the model is divided into three parts: i) an encoder; ii) a shared classifier; and, iii) a decoder to pro- duce pseudo-samples. The encoder is trained to generate embedding with the same distribution, regardless of tasks. For this, it models the embedding space as a multi-modal distribution using the Gaussian Mixture Model (GMM). When a new task needs to be learned, the model uses the decoder with the GMM to generate pseudo-samples that are merged with the samples of the cur- rent task. So, with this expanded data set, the model goes through a new learning phase in these three components. By doing this, the encoder encourages all samples to be grouped together in a single embedding region. As a result, the categorization task is simplified. It works as a form of normalization.

Variational Autoencoder (VAE) is naturally more resilient to CF (van de Ven et al., 2020). Due to this, VAE is used in Incremental Learning both to generate images and classification tasks. The choice of the prior distribution over the latent space is crucial to a VAE present accurate results (Egorov et al., 2021). However, in an Incremental Learning process, choosing this distribution is not trivial, because the data changes over a lifelong term. Egorov et al. (2021) propose to use the gradient in an end-to-end training to find a prior that is a mixture of current data distribution and old distributions. However, it also presents image quality problems produced by the decoder.

Another way used to generate examples of known classes is by using class prototypes. The model can remember the class since it serves as an anchor. A prototype is a representation of the samples of a class and is usually determined by averaging all available samples. This is the strategy that requires less storage memory overhead, however, it requires that all samples from a class are clustered and separated from other classes. In these scenarios, simply add Gaussian noise to the prototype to get an ephemeral collection that can be used together with the training set in each training session (Zhu et al., 2021b; Petit et al., 2023).

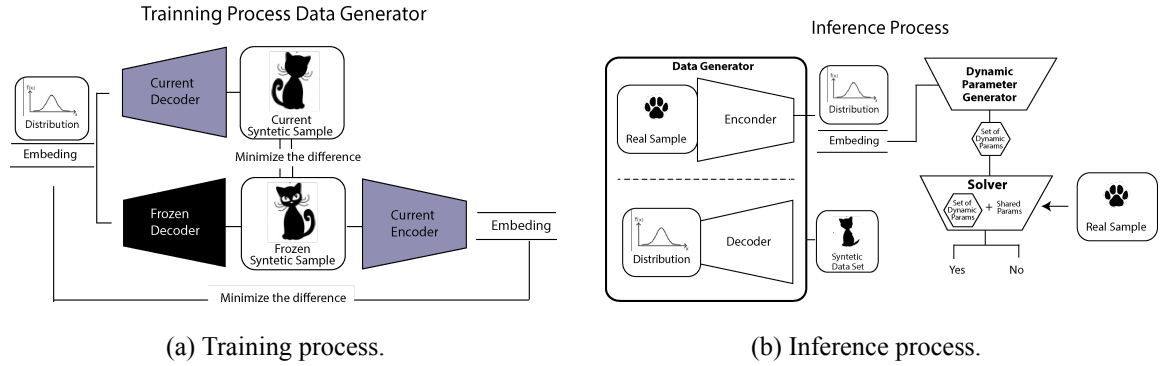

(a) Training process.   (b) Inference process.

Figure 4: Sequential training/infer of the PGMA framework. An image instance embedding is generated by the Encoder, then it is used by the DPG to create a set of parameters to adapt the Solver. The Solver is used to classify the image instance. The Decoder creates a synthetic data set to avoid CF when learning new classes.

Instead of forcing the approximation of samples in the embedding space, Hu et al. (2019) propose to adapt the value of the model's weights to handle instances from different classes. They did this by employing the model together with three auxiliary models: i) a *Solver*, which predicts the response of a task; ii) a *Dynamic Parameter Generator* (DPG), which creates a set of parameters to be merged in the Solver for each sample; iii) an *Encoder* that creates an embedding to be used as input by the DPG and the Decoder; and, iv) a *Decoder* which creates realistic data samples of tasks that are no longer available. This strategy allows mutating the weights based on the current processed sample. It helps the model to mitigate the semantic drift problem (Masana et al., 2020). This is also considered pseudo-rehearsal (and not a Dynamic Network) as the model maintains its structure but modifies its parameter values by utilizing data derived from prior knowledge.

Figure 4 shows how these models interact with each other during inference (left side) and train- ing (right side) time. Given an image of a new class, the Encoder generates an embedding vector. This vector is given to the DPG which outputs a set of weights. These weights are combined with those provided by the Solver to predict input class *y*, along with the samples of the new task. The models are trained with samples generated by the *Decoder*. *Data Generator*, a macro component composed of the *Encoder* and *Decoder*, can also suffer from CF. To mitigate this, the authors pro- pose to use a KD loss as illustrated on the right side of Figure 4. A frozen version of the last training session is used as a "teacher" in the current training session to remember the old behavior of the models. For this, a random embedding vector from a normal distribution is used to generate a synthetic image. Next, the objective is to minimize the difference between that synthetic image and the one generated by the current *Decoder*. In addition, the current *Encoder* must generate an embedding from the synthetic image and minimize the difference between its embedding and that used by the frozen *Decoder*.

While the results indicate strong performance in a two-task sequence, further evaluation is necessary to validate the method's efficacy in handling arbitrary task sequences and more complex tasks beyond MNIST. Future investigations should focus on expanding the evaluation to encompass di- verse scenarios, ensuring the method's scalability and robustness. Nevertheless, the use of synthetic data sets that are more similar to the current samples shows more promising results.

Hypernetworks are also used to generate values for the model's parameters (von Oswald et al., 2020) based on a task identifier. This approach passes the problem of CF to the hypernetwork. However, von Oswald et al. (2020) advocate this is less affected because the pair of input-output is less than these relations between data of classes from tasks. Although it still happens, they use a VAE to apply pseudo-rehearsal in the hypernetwork.

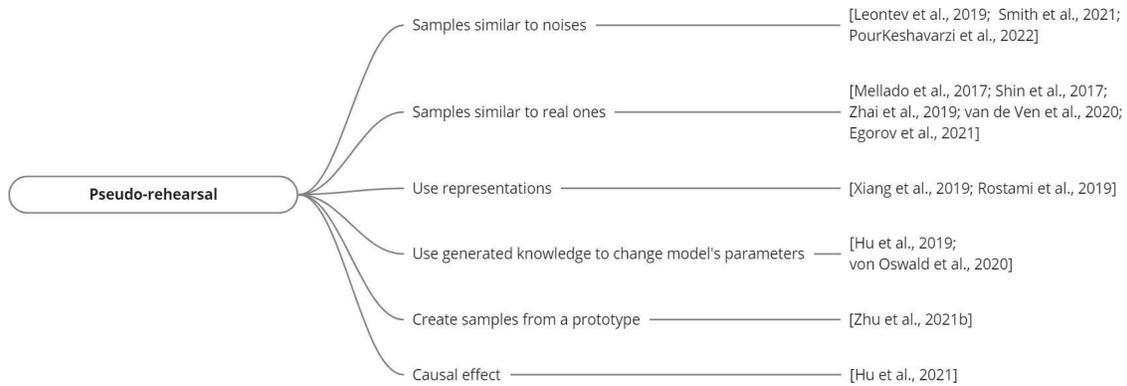

Figure 5: Overview of main approaches in the Pseudo-rehearsal category. In Pseudo-rehearsal, it does not require the storage of samples. Instead, the model leverages its existing knowledge to generate new samples as needed. The figure shows the main way to generate pseudo-samples and unusual ways to use these generated data.

Hu et al. (2021) explain the CF problem as a causal effect of the old data lost in new training. They argue that replaying a feature representation of data does not achieve the causal information of replaying the original data. However, they showed that it is possible to distill the colliding effect between the old and the new data on embedding space.

Other learning algorithms were also tested with pseudo-rehearsal. For instance, Contrastive Heb- bian Learning (Movellan, 1991) and No-Prop (Widrow et al., 2013) were tested by Hattori and Tsuboi (2018). However, they only tested on toy data sets and with shallow networks. Future work could perform tests with deep architectures and more data sets and also use GANs to generate syn- thetic data. Nevertheless, in our work, we focus on methods based on the gradient descent learning algorithm, so this work is out of our scope. Figure 5 depicts an overview of this category. It shows that the pseudo-examples can be used to change the parameter values of the model or distill the causal information. In addition, it shows the main ways to generate pseudo-samples. In the next section, we present works that allow for the storage of certain outdated data to prevent CF.

### 3.1.2 MINI-REHEARSAL

The idea behind mini-rehearsal is that a subset of the original data must be retained to prevent CF while training a model on a new task. The objective is to minimize the amount of old data that needs to be kept in order to allow the model to learn the new task without experiencing CF. This can be accomplished through selective data sampling. This subset is known as coreset. The main objective is to identify a representative subset of the training set. Therefore, with this subset stored, CF can be handled by the model.

The Greedy Sampler and Dumb Learner (GDumb), an established baseline of this category that demonstrates its strength, is a naive model that questions the advancements of contemporary approaches to address the CF suggested in the literature (Prabhu et al., 2020). A memory manager and a learner make up GDumb's two main modules. For each new class that is learned, a new bucket is made to hold its samples, according to the naive management practiced by the previous. The sample from the bucket containing the most samples is removed when memory is full. The learner uses samples stored in memory to train a DNN from scratch. Several studies that have been published in the literature are surpassed by this model.

There is a lot of research on how to improve the sample selection for the coreset (Shankar & Sarawagi, 2018; Hayes et al., 2019; Borsos et al., 2020; Liu et al., 2021; Yoon et al., 2022). It is possible to select the coreset that reduces the forgetting of knowledge using a bilevel optimization based on gradients (Borsos et al., 2020). However, it has a high computational cost despite the optimizations proposed by the authors. Using a GeForce GTX 1080 Ti GPU to select a coreset of 100 samples in a set of 1000, takes about 20 seconds, while selecting 400 samples in the same set takes more than three minutes. This shows that the time does not increase linearly. In addition, it only works with a static set of samples for the task, therefore it cannot be used in an Online Incremental Learning setup.

Most researchers advocate that the size of coreset has to be limited, but there are two ways to think about this: i) $M_c$ samples per class; or ii) $M$ samples to be managed to all classes. The former has the problem of total memory increase as the tasks learned increase, due to this, some researchers believe that compressing the data is a way to overcome this problem (Hayes et al., 2019; Wang et al., 2022a). Exemplar Streaming (ExStream) algorithm compresses the bucket data and is compared to other two compression algorithms: i) Online k-means; ii) and, CluStream; and two replacement algorithms: i) reservoir sampling ii) and, first-in-first-out queue. The results showed that when we have a bucket size above 256, all algorithms behaved similarly. However, when the bucket size is small, like four or eight, the correct choice of samples can improve the results. For instance, ExStrem showed an improvement of more than 50%, in some cases, compared to the reservoir algorithm (Hayes et al., 2019).

The sample compression causes a tradeoff between quality and quantity, however, it is possible to have gained in accuracy, even with lower quality samples (Wang et al., 2022a). It has also been noted in the literature to apply mutations to coreset samples in order to maximize memory retention. Mnemonics (Liu et al., 2020a) consider each bucket as a variable and apply a post-training phase, where the samples stored in the memory mutate their values to be considered more representative. Despite it being more computationally efficient than the above methods, it still requires to be done in the offline phase. Jin et al. (2021) apply gradient-based editing on samples from coreset (in an online fashion), instead of replacing them, however, they were looking to unbounded task incremental learning setup, where the data distribution changes without a sense of task identifier.

Another way to reduce the amount of memory required is to consider storing the representations (feature maps) instead of raw samples to perform the rehearsal process. REMIND (Hayes et al., 2020) showed that it does not present better results in scenarios where each task is learned sequentially, however in scenarios where classes came in any order (online learning), it presents state-of-the-art results. To achieve this, they used the MixUp (Liang et al., 2018) technique in the stored samples to get more robust and representative sets in the replay process.

Given that the coreset has a fixed size of $M$, we need a method to replace, update, or remove samples from the classes' bucket when memory is full. Applying the herding selection algorithm, ICaRL (Rebuffi et al., 2017), detailed in Section 3.5 as it is a hybrid technique, was a pioneer in this field. Herding selection chooses the samples closest to the centroid of the samples of each class. Reservoir sampling is the most used strategy to manage the coreset. This is a naïve strategy that gives each sample the same chance of being selected to be removed from the coreset once it is full. As a result, it cannot be assumed that the coreset will always be balanced, however, empirical research suggests an approximation of this (Hayes et al., 2019).

Metrics can be employed to determine the selection of samples that will be retained within the coreset. Shapley value was used to estimate the individual contribution of each sample to the performance of the model (Shim et al., 2021). Maximally Interfered Retrieval (MIR) uses the loss to check if the incoming sample is more negatively impacted by updating the model parameters than the already stored data. Information theory was used to manage the most relevant samples to keep in the coreset using the Bayesian model to compute the criteria efficiently by exploiting rank-one matrix structures (Sun et al., 2022). Classifier's confidence is another metric that was used in the literature. It is possible to select the elements with the largest or worst confidence. In the former case, it is selecting samples that the model has no problem hitting, while the latter are samples that the model has difficulty inferring correctly. According to empirical studies, choosing the most trustworthy ones produces better results (He et al., 2018).

An RL agent was trained to decide what proportion of memory should be allocated to each class and also which samples it should keep (Liu et al., 2021). As the agent can also suffer from CF, it is only trained on the first task, splitting the data into $N$ groups to simulate various tasks. In an Online Incremental Learning setup, the problem of an imbalanced and noisy data set is more predominant. Online Coreset Selection (Yoon et al., 2022) manage the memory following three selection objectives: i) minibatch similarity, to select representative samples to the current task; ii) cross-batch diversity, to reduce the redundancy among the samples of the current task; and, iii) coreset affinity, to minimize the interference between the selected samples and knowledge of the previous tasks.

All pretrained models can use the coreset in an Incremental Learning scenario, transforming it into a Memory Augmented Neural Network (MANN) (Santoro et al., 2016). The memory of MANN is a table in which each row is a tuple $\{l_m, v_m, \alpha_m\}$, where $l_m$ is the instance label, $v_m$ is the embedding generated by the last layer of the model (just before softmax activation) and $\alpha_m$ is the weight attached to the row. During inference, they interpolate the prediction of the pretrained model and prediction of the memory, using the top $N$ most similar tuples comparing $v_m$ with the embedding generated by the last layer's model. These two values are interpolated by means of a set of weights, which is generated by a recurrent neural network (RNN) in each prediction.

At training time, the truth label is used to update the memory if a mistake is done by the model. In addition, only tuples where $l_m$ is equal to the truth label are updated, so that rare classes are not forgotten. The RNN is also trained in an online fashion to decide the relevance of the prediction done by the memory and the pretrained model. Although the idea of the work implies that CF is avoided, there are no evaluations in this regard. Instead, the authors only evaluate the forward trans- fer knowledge without checking if older classes are still remembered by the model. Nevertheless, it is evident that this memory aids in addressing the issue of imbalanced data sets by leveraging the top $N$ similar tuples (Shankar & Sarawagi, 2018).

Following the approach based on feature embeddings (last layer's output), Sprechmann et al.

(2018) propose the adoption of a hash table using the embedding as key and the truth label as value. The model is divided into three components: i) the embedding generator; ii) the hash table memory; and, iii) the classifier. During training time, all models are updated. In the hash table, samples are appended as a circular buffer, while the embedding generator and the classifier are updated with backpropagation. In inference time, a K-Nearest Neighbors (KNN) is applied to the hash table keys, and the result is used to modify the classifier weights on demand. This is an online adaptation, as done by Hu et al. (2019), however, instead of using a model to generate the drift in the weights, a little training session is performed to find this $\Delta$ drift. As the memory has a fixed size and acts as a circular buffer, the model has a tendency on the most recently trained classes, making it difficult to remember rare classes or old tasks.

The literature makes numerous and varied uses of the coreset to reduce the CF. MeRGAN (Wu et al., 2018) analyzes two ways: i) joint training with replay; and ii) replay alignment. In the first, the samples are generated conditioned in the label of past tasks to be used together with the samples of the current task. While in the second, it is the application of KD in the Generator model, making the generator of the current task a student of the generator learned in the last task. Gradient Episodic Memory (GEM) (Lopez-Paz & Ranzato, 2017) uses it to add a hard constraint in the updating step of the model. It only accepts changes in the parameters of the model if the error on coreset does not increase. This avoids forgetting in the coreset and allows backward knowledge transfer. However, the general error in the class can still increase. Moreover, the model can lose its ability to learn new tasks due to this hard restriction. Chaudhry et al. (2019a) proposes to relax this constraint and start to accept updates despite increasing the error by a threshold. They called this method A-GEM. It outperforms GEM in accuracy and in the number of tasks that the model can learn.

To regularize the training in a different approach, Tang et al. (2021) divide the gradient into two pieces and use the coreset to do this: i) relative to all tasks; and, ii) relative to the specific task. Then, they apply one constraint to each part. In the first, the gradients have to be in the same direction. In the last, they have to be orthogonal to another task-specific gradient from other tasks. Another regularization method adopted was the Elastic Weight Consolidation (Kirkpatrick et al., 2017), used in the Natural Language Generation (NLG) domain (Mi et al., 2020). This constraint is a technique from the Sub-networks category that will be discussed in Section 3.3.

Hou et al. (2019) and Wu et al. (2019) hypothesized three main reasons why methods that use mini-rehearsal techniques present in the problem of forgetting: i) imbalanced magnitudes; ii) semantic drift; and, iii) ambiguities. Hou et al. (2019) create LUCIR to prevent these three points. For the former reason, they noticed that the value of logits referring to classes of the current task tends to have a magnitude much higher than that of the classes of old tasks due to data imbalance, where the samples of the classes of the current task are abundant while the samples of the old classes are scarce. Therefore, they apply cosine normalization to the logits to make them all of the same magnitude. To face the semantic drift, they use KD in the embedding layer. Finally, to handle ambiguities among the classes, they use a margin ranking loss.

BiC appends a final layer to the model responsible to equalize the bias of the logits in relation to the new classes (Wu et al., 2019). For this, the samples of the current task are divided into two sets, validation, and training. The training set, which is still abundant, is used to train the model down to the logits layer, this is called stage 1 of the training session. Then, the validation set, which is balanced against the in-memory set, is used to fine-tune the last linear layer, which is used to make predictions. In the first stage, they also use KD to prevent semantic drift. They advocate that, in the second stage, the fact of training in a balanced data set, the model can handle the problem of ambiguities by itself. BiC is not categorized in dynamic networks because despite it appending a final layer on the model, it only happens on the model's creation, not in its lifelong term.

The problem of imbalanced magnitudes was also observed by Belouadah and Popescu (2020). To handle this issue, they use extra memory to store the statistics of classifier nodes of classes in the current training session when the data of these classes are abundant. In the next training sessions, when the data of these classes are sparse, the magnitude of their nodes tends to drop. Therefore, these statistics are used to rescale the weights of those parameters. Zhang et al. (2022) demonstrates that in an Online Incremental Learning setup, employing random transformations on the coreset to increase its size helps the model generalize knowledge and mitigate CF. Their work serves as a strong baseline for this particular setup.

Lee et al. (2019) propose to use a stream of unlabeled data from the internet to overcome the problem of unbalanced data. They advocate that this alleviates the three problems above. Moreover, they indicate this approach is better than fine-tuning the model in a post-training with a restricted amount of data. The data that comes from the stream are labeled by the model itself versioned on the last training session, in such a way as to distill the knowledge of old classes. In the context of object detection, Dong et al. (2021) use unlabeled data to get a sampling of objects that are not presented in the current task anymore. Moreover, they use a scheme of dual networks to apply KD of this old knowledge.

Ahn et al. (2021) argue that this bias is mainly caused by computing the softmax probabilities by combining the output scores for all the old and new classes. Thus, it is proposed to separate the softmax layer, leaving a head for each task and applying KD with the coreset during training. However, Zhu et al. (2021a) show that the models also create a bias in the embedding space for the most recent classes. To address this, it is proposed to perform a semantic augmentation (semanAug) in the latent space with the old samples.

Belouadah et al. (2020) go in this direction and defend that forgetting mainly affects the classification layer. They proposed that vanilla fine-tuning can mitigate CF since classifiers learned in old tasks can be kept to standardize the weights against all classifiers learned in each task. They argue that this is necessary due to the bias towards the new classes. To make the weights fairer after learning a new class, the standardization of initial weights is done.

Hou et al. (2018) propose the Adaptation by Distillation (AD) method to show the advantages of KD in terms of Incremental Learning. In the training process, the lifelong term model does not directly learn from the truth labels, instead, an auxiliary model, called "*expert*", is trained with original data. After that, the main model is trained by this expert, using its soft labels. They showed that this process gives the model a more generalized knowledge, reducing the effects of CF. Despite the KD being widely used to mitigate CF, Belouadah and Popescu (2019) show, in their experiments, that it is worst to model accuracy when the model has access to a representative coreset.

The Meta-learning approach was proposed to create a model agnostic (Rajasegaran et al., 2020). This model is not a specialist in any task; however, its solution is near all tasks. Therefore, it can use the coreset with the one-shot learning strategy to solve any task in its lifelong term. Von Oswald et al. (2021) find that sparse learning emerges, due to a large fraction of learning rates dropping to zero in this type of model (MAML). They showed that it can be used to focus the learning on the most appropriate regions of the model. Perez-Rua et al. (2020) show that this approach also works in the context of object detection; and, XtarNet (Yoon et al., 2020) uses meta-learning to train a MetaCNN to produce an embedding to new classes. These new embeddings are combined with the ones generated by a pretrained network. Then, the combined embedding is used for the final classification. Part of the data from each class is maintained in memory to be used as a query set and support set in meta-learning.

Generative classifiers are less prone to CF because the decision boundaries of *n* classes learned sequentially are the same as when trained together (Banayeeanzade et al., 2021). GeMCL extends OML (Javed & White, 2019) replacing its discriminative classifier with a generative one (Bayesian classifier). Henning et al. (2021) propose a Bayesian framework where task-conditioned posterior parameter distributions are continually learned and compressed in a Hypernetwork. They reported almost no forgetting and show it scales to modern architectures such as ResNets.

Recently, with the increasing attention on Transformers, researchers have begun investigating their potential benefits in addressing the issue of CF (Wang et al., 2022b, 2022c). They propose the hypothesis that by identifying the appropriate Prompt, the model can effectively prevent CF. However, a challenge arises in determining the ideal Prompt for each instance. In their study, Wang et al. (2022b) propose a method to learn and store the Prompts associated with each class within the coreset. Each Prompt is linked to a key, which is the embedding of a sample from the corresponding class. During inference, the current sample's embedding is utilized to select the most suitable Prompt. This approach opens up a new research direction, as it shifts the focus from CF avoidance to discovering the optimal Prompt.

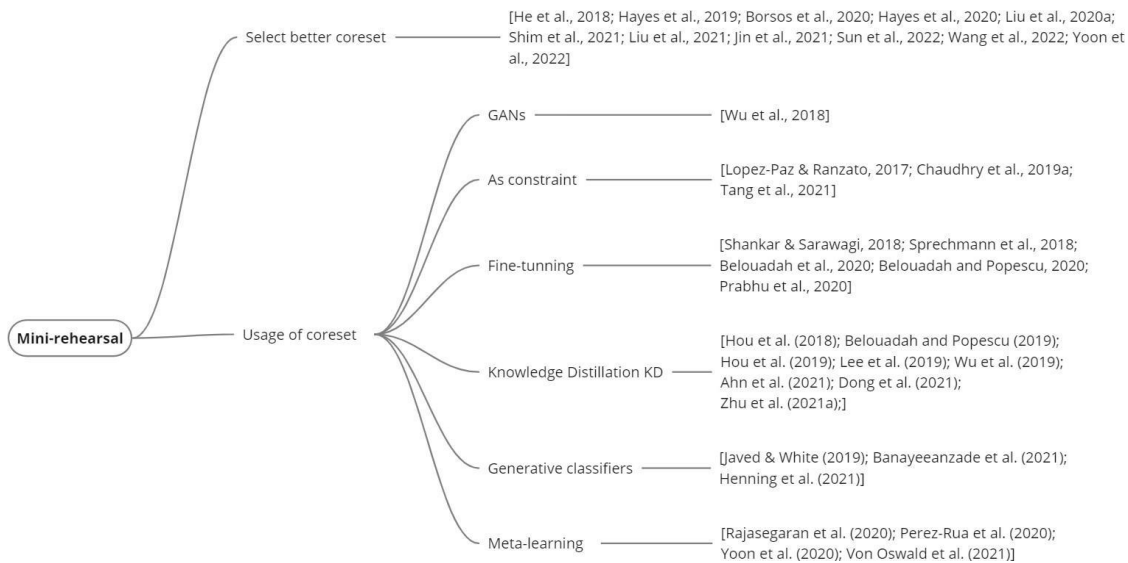

Figure 6: Overview of main approaches in Mini-rehearsal category. In Mini-rehearsal the model can store a limited part of samples from old tasks. Existing works that use these data to avoid CF and others that research better ways to select the coreset.

Figure 6 depicts an overview of this category. Although rehearsal methods show good results, they present an inherent problem of overfitting in the coreset (Chaudhry et al., 2019b). Verwimp

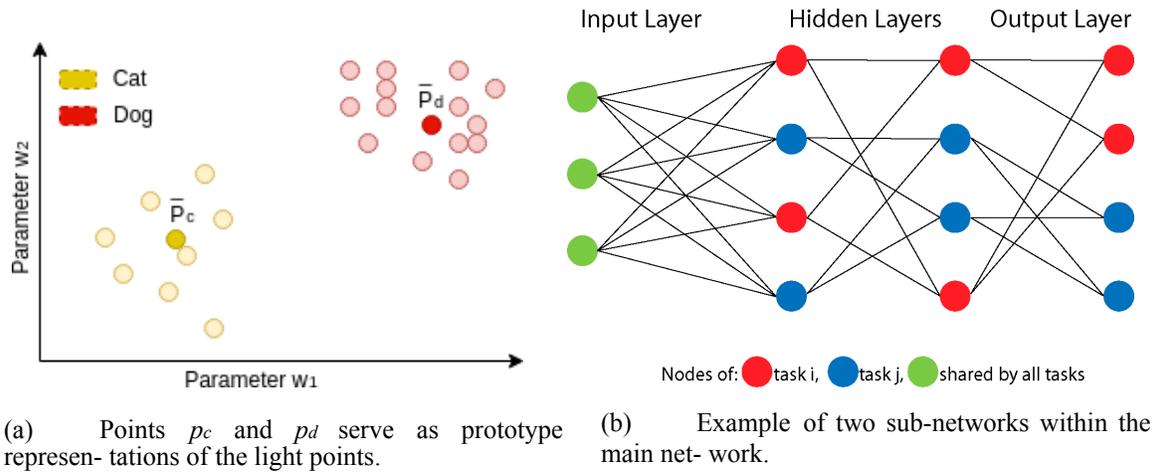

(a) Points $p_c$ and $p_d$ serve as prototype representations of the light points.

(b) Example of two sub-networks within the main net- work.

Figure 7: (a) A prototype representation condensing the entire class into a single point, reducing the impact of imbalanced data among the classes. (b) A shared sub-network approach.

et al. (2021) advocate that this result is observed because the models that use coreset stay in the same low-loss region after a task has finished. However, they point out that more research in this direction is required. There are many spaces in research to evaluate the best approach to select the best samples to maintain in memory. In the next section, we present the works that use techniques focused on learning representations, that we classify as Distance-based methods.

### 3.2 Distance-based

Within the **Distance-based** category, Rebuffi et al. (2017) propose two primary approaches: i) fixed data representation; and ii) learning representation. In the former, a fixed embedding generator is used in every Incremental Learning step and the major challenge is to ensure that the generated embeddings are representative of all upcoming classes in a way that keeps embeddings from samples belonging to the same class grouped together but apart from embeddings from samples of other classes. In the latter, the embedding generator also suffers adjustment in its parameters in each training step. The main challenge is the tracking of the embedding region of the classes when the embedding generator updates and the model does not have access to samples from those classes.

The use of prototype classes is a common technique employed in many methods within the **Distance-based** category. A prototype class is an embedding that represents the central tendency of other samples from the same class. Figure 7a provides an example where the embeddings are colored light yellow and light red, representing cats and dogs, respectively. The dark yellow point ($p_c$) is the prototype class of cats and is created by averaging the features of the embedding of the cats. Similarly, the dark red point ($p_d$) is the prototype class of dogs and is also created by averaging the features of its class.

The use of a prototype class, instead of all samples, helps Incremental Learning to reduce memory requirements and avoid training with classes with unbalanced sample amounts. The first problem is handled because the model has to store only one data point. The second problem was noticed by Belouadah and Popescu (2019). They showed that the models suffer from a bias in favor of new classes. This bias is caused by the number of samples of new classes in relation to the old ones.

The choice of embedding representation is crucial in determining how well a classifier can avoid forgetting. When the representations of samples from different tasks are aligned in embedding space, learning one task facilitates the learning of the other. Conversely, when they are orthogonal, learning one task does not affect the learning of the other. In a recent study, Javed and White (2019) used a meta-learning algorithm to train a feature extractor that searches for representations that maximize parallel or orthogonal embeddings, thus enabling the classifier to determine a region for each task's classes.

In a few-shot learning setting, aligning visual embedding and word embedding is another technique for choosing the space for each class in the embedding space. Cheraghian et al. (2021) pro- posed a method where the model takes an image as input and produces an embedding that should be close to the word embedding that represents the class, instead of generating a label as output. They tested this method with Word2Vec embedding (Church, 2017) and GloVe embedding (Pennington et al., 2014). By doing so, a model that can handle natural language classification problems can be used without experiencing CF.

In situations where the embedding generator updates its weights during training, the embedding space of a class can change. To address this issue, Semantic Drift Compensation (SDC) (Yu et al., 2020) updates all prototypes at the end of each training session to discover the new region in the embedding space where a class will be allocated. To calculate this change, the model determines the position of the incoming data in the embedding space before training and again at the end of the session. The difference between these positions is then used to update the prototypes. However, it's important to note that SDC assumes that all class embeddings continue to be clustered with their respective prototypes, which is not guaranteed. Therefore, while this technique can be effective in certain scenarios, it may not always provide accurate updates to the prototypes.

Incremental Unsupervised Learning of representations using the MixUp technique (Liang et al., 2018) has yielded promising results that surpass those obtained with Supervised Incremental Learn- ing. One possible explanation for this phenomenon is that unsupervised learning generates a smooth loss landscape (Madaan et al., 2022). This approach has been tested with Barlow Twins (Zbontar et al., 2021) and SimSiam (Chen & He, 2021). The main classifier used in this category is based on cosine similarity. Two improvements have been proposed for this classifier: (i) generating multiple perspectives of each sample, and (ii) using a Contrastive loss. In the first approach, each sam- ple is used to generate multiple views of itself by applying rotations. In the second approach, the generated samples are used as anchors in a Contrastive loss (Wu et al., 2021).

When a model is trained using the Joint Training approach, where all classes are trained to- gether, it leads to a better decision boundary within the embedding space compared to training in an incremental fashion. In general, this approach tends to produce a more robust class separation, effectively enhancing the model's ability to distinguish between different classes and improving overall generalization performance. Therefore, Shi et al. (2022) propose an objective function that aims to enforce the decision boundary generated through incremental training to mimic that of Joint Training.

Figure 8 provides an overview of the Distance-based category. One of its advantages over other categories is that it can expand its known classes without changing its structure, focusing only on the embedding space. However, it remains unclear how many classes an embedding space can effectively support without causing confusion. In the next section, we will introduce a category that attempts to address this issue by identifying sub-modules within the model that can mitigate CF.

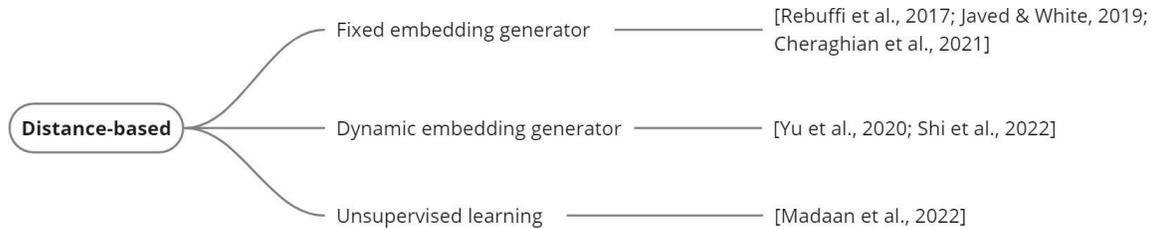

Figure 8: Overview of main approaches in Distance-based category. In Distance-based, cosine classifiers are the most used. It assumes that the embeddings of the same class are clustered in the same region.

### 3.3 Sub-networks

When learning a new task, previously learned patterns are replaced by the current training patterns, which can lead to forgetting. For example, a model may use a neuron connection as a positive stim- ulus to improve its performance when learning its initial task, and later, when learning a second task, that same neuron may need to be used as a negative stimulus, which can decrease the model's per- formance in the previous task. To address this issue, some approaches try to encode the knowledge needed for each task in separate sections of the model with minimal overlap, as shown in Figure 7b. By considering a neural network as a high-capacity model with redundant connections, it becomes feasible to selectively update specific subsets of neurons during the training of each new task. These subsets, known as Sub-networks, consist of distinct sets of neurons within the main model, and the aim is to minimize their overlap.

Goodfellow et al. (2014a) led a pioneering study on the subject of CF. They conducted a series of experiments with various activation functions and regularization techniques to determine which combinations were less prone to CF. They found that the combination of the ReLU (Nair & Hinton, 2010) activation function and Dropout (Hinton et al., 2012) regularization was the most effective. Dropout works by randomly disabling a subset of connections during each training iteration, which forces the model to learn using the remaining active connections. This encourages patterns to be encoded redundancy in different subsets of connections, meaning that even if some connections change in future tasks, there may still be alternative paths using different subsets of connections that can solve the initial task. This suggests that a model can be composed of multiple sub-models. Therefore, choosing a good training regimen can reduce CF in a wide range of models (Mirzadeh et al., 2020), such as the best combination of dropout, weight decay, learning rate, batch size, and optimizer. It's important to note that these training regimens can also be used in conjunction with other techniques discussed in this survey.

The Sub-network category can be classified into two types: Soft Sub-networks and Hard Sub-networks. Hard Sub-network-based approaches aim to identify a subset of parameters that represent the learned knowledge of a specific task and then freeze them. Consequently, when the model has to learn a new task, it adjusts a different subset of neurons to form small networks within the model. This produces a submodel for each learned task, each with varying performance. However, because a new subset of parameters must be frozen for each trained task, the number of tasks the model can learn is limited, as the overall network size remains constant.

ALEIXO, COLONNA, CRISTO & FERNANDES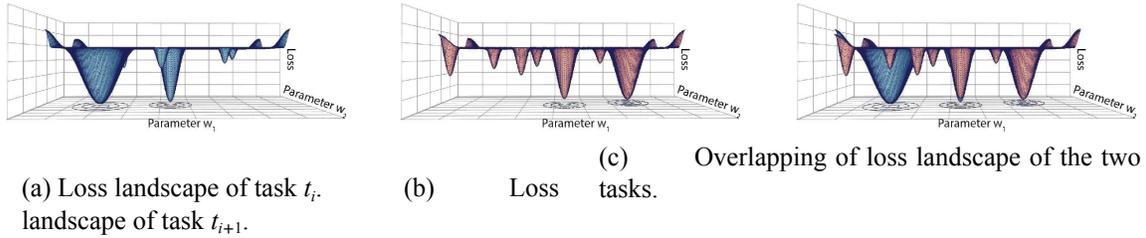

(a) Loss landscape of task $t_i$.
(b) Loss landscape of task $t_{i+1}$.
(c) Overlapping of loss landscape of the two tasks.

Figure 9: Loss landscape of two tasks. Each task has more than one possible value to its weights (Parameter $w_1$ and Parameter $w_2$) that shows similar results in the final mean accuracy of the model. On the right side (c), we observe the two loss landscapes overlapping, and the star point represents a region to weights that is a good solution to both tasks.

Like Hard Sub-networks, Soft Sub-networks also select a set of important parameters for task $t_i$. However, instead of keeping these parameters frozen by default, a term is added to the loss function that penalizes changes to the parameters of task $t_i$. This ensures that the network will do its best to avoid changing the important parameters of task $t_i$ while generating values for the parameters of task $t_{i+1}$. In this way, Soft Sub-networks try to overcome the capacity limitations observed in Hard Sub-networks approaches by allowing neurons to suffer little changes during training new tasks.

Adding a penalization term to the loss function changes the loss landscape. It has been established that in the loss landscape, multiple local minima can yield satisfactory levels of accuracy for a specific task, as illustrated in Figure 9 (Lee et al., 2017a). By adding a penalty term, each submodel can be forced to reach a local minimum that provides optimal performance for each task simultaneously. With Soft Sub-networks, different local minima can correspond to different solvable tasks for each submodel. Therefore, the goal of successful Soft Sub-network training is to bring the local minima of each submodel closer together, allowing the parameters of one submodel to be reused by another submodel. For example, after learning task $t_i$ and starting to learn task $t_{i+1}$, the network is forced to find an acceptable solution for task $t_{i+1}$ that is as close as possible to the solution found for task $t_i$. The works that most contributed to these two approaches will be discussed in the following subsections.

### 3.3.1 SOFT SUB-NETWORKS

In the Soft Sub-networks, each submodel specializes in solving a task that shares parameters (weights) with the other submodels, which helps to avoid the capacity limitations of Hard Sub-networks. However, the subset of important parameters for task $t_i$, which will be shared by the submodels of subsequent tasks $t_{i+1}$, $t_{i+2}$, . . ., cannot go through a lot of changes when those tasks are being trained. To keep these parameters useful for the original task, they must be identified prior to beginning the training of task $t_{i+j}$, where $j$ represents a positive integer value. The Fisher Information Matrix (FIM) was used by Kirkpatrick et al. (2017) to determine which parameters are most important for solving task $t_i$. This information is then incorporated into the loss function as a regularization term to prevent changes to these parameters during the training of task $t_{i+1}$. This method is called Elastic Weight Consolidation (EWC) and is one of the most widely cited approaches in the field of Soft Sub-networks.

Hong et al. (2019) introduced Predictive EWC as an enhancement to the existing EWC algorithm. In Predictive EWC, before the model starts learning a new task $t_{i+1}$, it initially processes the new training data using the current model. This preprocessing step enables the model to identify and discard samples that it can handle accurately, retaining only those with lower accuracy. Con- sequently, a refined data set is formed, based on the assumption that certain samples are no longer necessary for the model since it has already learned how to handle them effectively. This refinement of the data set leads to a reduction in training time.

In a similar vein, Kobayashi (2018) suggests that when data from different tasks exhibit substantial distribution differences, a more aggressive regularization approach should be applied to important parameters. To address this concern, the author proposes a conditional loss called Check Regularization, which facilitates parameter grouping based on each task, while still allowing parameter sharing between similar tasks. By incorporating Check Regularization, the model can adapt its regularization strategy to account for variations in task distributions, ultimately improving per- formance across diverse tasks.

The EWC algorithm assumes that the FIM is a diagonal matrix. However, in practice, this as- sumption does not hold true. To address this limitation, a technique proposed by Liu et al. (2018) can be employed. The idea is to rotate the model's parameters in such a way that: i) the model's output remains unchanged; and, ii) the FIM calculated from gradients becomes approximately di- agonal. This rotation-based approach has demonstrated superior accuracy compared to the original EWC method. However, it comes at the cost of increased processing time.

While these approaches have shown improvements, the use of strong regularization techniques can potentially hinder the model's ability to acquire new knowledge. Additionally, there are two no- table drawbacks associated with these methods. Firstly, they do not consider the presence of batch normalization layers, which are commonly used in DNN. Secondly, they involve a high computational cost during post-processing when calculating the FIM. These drawbacks should be taken into consideration when applying these techniques in practical scenarios.

In CF research, the influence of batch normalization layers is often overlooked due to their potential to degrade model accuracy when the data distribution changes over time, rather than im- proving it. However, to tackle this particular drawback, a promising solution called the Continual Normalization (CN) layer has been proposed by Pham et al. (2022). The CN layer performs spatial normalization on the feature map using group normalization. By incorporating CN layers into CF models, the adverse effects of batch normalization on changing data distributions can be mitigated, thereby preserving model accuracy over time.

To overcome the high computational cost associated with calculating the Fisher Information Matrix (FIM) during post-processing, the Synaptic Intelligence (SI) technique offers a viable solu- tion. SI involves the creation of a matrix $\Omega_k$ that represents the importance of each parameter. This matrix is calculated during the training session by applying a small perturbation to the parameter values and observing the resulting degradation in the solution, as described by Zenke et al. (2017). Parameters that have a more negative impact are considered more important.

Similarly, Jung et al. (2020) also employs an importance matrix within the SI framework, but with two main differences. Firstly, instead of calculating the matrix during training, they compute it during inference time. This shift in computation reduces the post-processing computational burden. Secondly, they nullify the output weights of nodes deemed unimportant. When the model needs to learn a new task, these nullified weights are randomly reinitialized. This approach provides the model with an opportunity to acquire new and different knowledge. By nullifying and reinitializing the weights, the model can adapt its architecture and synaptic connections to accommodate the requirements of the new task.

Another approach to address the computational cost during post-processing involves considering all parameters as important and striving to maintain them within the same region defined by the first task. This can be achieved through the utilization of the Monte Carlo Variational Inference method (Blundell et al., 2015). By minimizing the variation of all model parameters using the Kullback-Leibler divergence, the model seeks to preserve their values across different tasks (Nguyen et al., 2018).

To further refine this approach, Ritter et al. (2018) introduced the concept of online Laplace Approximation with Kronecker. By leveraging this technique, they estimated the posterior distribution of the parameters. This estimation not only considers the correlation between a parameter and itself in the next task but also accounts for the correlation with other parameters. By incorporating these correlations, the model can capture more nuanced dependencies between parameters, resulting in a more accurate estimation of the posterior.

Aljundi et al. (2018) advocate that some things need to be forgotten for new knowledge to be learned. They maintain a matrix $\Omega$ during inference time. This matrix is updated by each inference, observing which parameters are more activated when the test data is processed by the model. In this sense, the model gives more importance to connections that are more used in test time. This method was called Memory Aware Synapses (MAS). The principal problem of MAS is that it tends to forget rare classes quickly. Wang et al. (2021) show that this selective forgetting is consistent with the underlying mechanism of biological forgetting. Thus, they add a term to calculate the weight importance matrix to decide which parameters, despite being important for the previous task, should be modified to maximize the probability of the new task being learned.

To incorporate a temporal factor in the strength of synapses, the Benna-Fusi model proposed by Benna and Fusi (2016) was integrated into a DNN. This model introduces the concept of gradu- ally changing synapse strength over time, resulting in some synapses being reinforced while others are lost. This temporal modulation of synapse strength enables the network to adapt and incorpo- rate new information while retaining important knowledge from previous tasks. Although initially tested in the context of RL, where there is a natural notion of time due to the sequential nature of agent actions, the concept of temporal modulation can be extended to other domains as well. By incorporating the Benna-Fusi model into a DNN, researchers can introduce a temporal dimension to the learning process, allowing the network to dynamically adjust its synapse strengths based on the sequence of tasks encountered.

There are also alternative strategies for the creation of Soft Sub-networks, apart from penalizing critical weights. To eliminate the module of parameter importance, a technique called Incremental Moment Matching (IMM) proposes to train a set of models $M_{t^i}$, all of them with the same structure. Each model is addressed to learn a unique task. In the end, all models are averaged and the resulting model should be able to solve all tasks in $T$ (Lee et al., 2017a). Tasks are not learned in parallel, because for this strategy to work, the initial weights of the model $M_{t^{i+1}}$ need to be cloned from the already trained model $M_{t^i}$. In addition, $L2$ regularization is applied to the weights of models $M_{t^i}$ and $M_{t^{i+1}}$ to ensure that the parameters of the solution of the second task are close to the parameters of solution of the first task.

Another alternative to solve the selection problem on Soft Sub-networks is to try to select a subset of parameters to solve each task based on context. Generally, the input distribution for each task is regarded as the task's context. Orthogonal Weights Modification (OWM) uses the input distribution to choose the filters used by the model to solve the tasks. It is supported by finding orthogonal filters for each task and, in inference time, using the task identifier to retrieve the right filters (Zeng et al., 2019). To achieve this, when a new training session happens, the parameters can only be modified in the direction orthogonal to the subspace spanned by previous tasks. Ke et al. (2021) use two RNNs to define the context: i) the main; and; ii) the auxiliary. The latter is trained to choose the optimal parameters of the former based on input distribution. The goal of the auxiliary network consists to find a subset of connections on the main network that best solves the current task without training. As a result, this subset is also used for similar tasks, and instead of forgetting happening, it improves the quality of representations. The main network is responsible to aggregate the knowledge of all tasks in their parameters. And, use the subset parameters selected by the secondary network to train in the current task with supervised learning, nullifying other parameters. Inspired from Learn without Forgetting (LwF) - discussed in the next section -, Learn without Memorizing (LwM) uses Knowledge Distillation (KD) (Hinton et al., 2015) to alleviate the forget- ting in training sessions (Dhar et al., 2019). While LwF is focused on Task Incremental Learning setup, the LwM is focused on Class Incremental Learning setup. This means that it does not create an entire classifier for each task, only appends one output for each new class in the classifier that already exists, and regularizes the internal paths of the model to classify each new task. To achieve this, KD is applied to the last two layers of the model: the embedding layer and the classification layer. KD is a technique where a smaller, student network learns from a larger, teacher network, to improve its performance. However, in Incremental Learning, the teacher is usually the model with weights frozen in the oldest learned task.

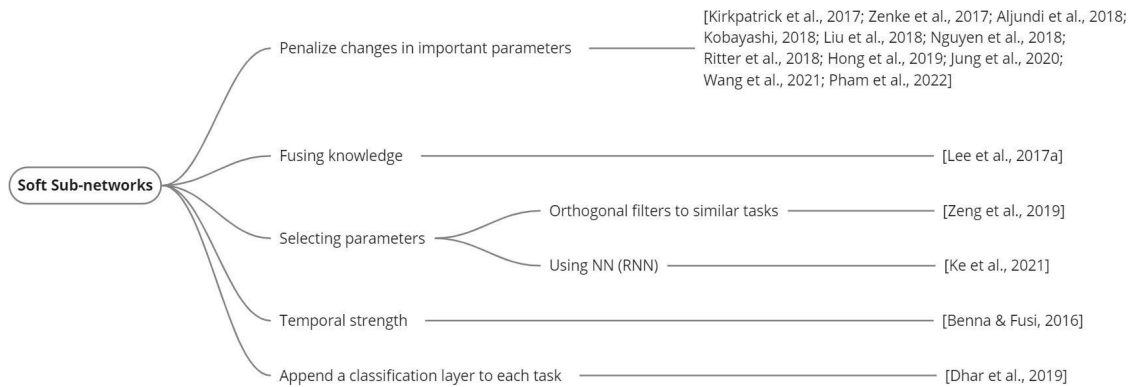

Figure 10: Overview of main approaches in Soft sub-networks category. In Soft sub-networks, the tasks can share part of the selected parameters to solve other tasks. The works discussed in this section are clustered by their approaches.

Most of the research in this category focuses on penalizing or controlling parameter changes to increase model stability, but this can reduce plasticity, making it difficult for models to learn new tasks. While preserving prior knowledge is important, it's essential to balance stability and adaptability to efficiently acquire new knowledge while retaining what has already been learned. Figure 10 provides an overview of this category. One challenge with these methods is that there is often a drift in the distribution of input data between tasks. If the tasks are similar, such as the MNIST splits, all methods in this group can handle the problem because the drift is low, and the samples belong to the same domain. The choice of which method to use in this group depends on training time, with Synaptic Intelligence (SI) or Incremental Moment Matching (IMM) being a better choice for shorter training times. However, when the tasks are from different domains, the drift is high, and none of these methods can handle the problem. This occurs because solving task $t_{i+1}$ requires changing critical parameters that were learned to solve task $t_i$. To address this problem, another research approach, known as Hard Sub-networks, was proposed. In Hard Sub-networks, the connections used to solve task $t_i$ are frozen, and this approach will be discussed in the next section.

### 3.3.2 Hard Sub-networks

In the Hard Sub-networks category, the main objective is to find a set of weights inside the main model. However, different from the Soft Sub-networks category, after the set of important weights was selected for a task $t_i$, they cannot be changed in the training session of tasks $t_{i+j}$, where $j$ represents a positive integer value. Coop et al. (2013) led a pioneering study on this category proposing the Fixed Expansion Layer (FEL) Networks. FEL expands the original neural network adding a layer after each hidden layer. These additional layers are larger than their predecessor hidden layer and each neuron of the original layer is connected only to a subset of neurons in the new hidden layer. In addition, the weights of these new layers are frozen, thus they are never changed during the training phases. This expansion of the layers helps the model to avoid the overlap of knowledge over tasks. As the choice of connections between the hidden and the additional layer is random, they use an ensemble of FEL networks. Therefore, this technique is not categorized as a hybrid or dynamic because it only changes the model's structure only at the beginning.

The selection of a subset of input neurons to combat the CF was introduced by Goodrich and Arel (2014). The proposed selection forces different regions of the input space to be learned in different nodes avoiding the knowledge overlap. Gepperth et al. (2015) propose to use a Self-Organized Map (SOM) to correlate the input with the neurons. In each training session, SOM is updated if the model fails, otherwise, just the last layer suffers fine-tuning. Therefore, for each task (represented by the statistical distribution discovered by SOM) a new subset of parameters is selected. Similar work was done by Lancewicki et al. (2015); however, instead of using SOM, they used the unsu- pervised algorithm *KNN*. To separate neurons they used Mahalanobis distance with a sample and sparse covariance matrix generated by a Shrinkage estimator (Ledoit et al., 2012). Other researchers concentrate on choosing parameters from the full model due to the limited resources in the input layer.

A Genetic Algorithm (GA) approach was used by Fernando et al. (2017) to find the best minimal subset of parameters to learn each task sequentially. When the training starts, each chromosome is a subset of parameters. The model freezes all other parameters and applies a traditional gradient learning mechanism. When the solution gets stable, the model reserves these parameters for this task. Thus, when a new task must be learned, it can only use the set of non-reserved parameters. It is important to consider solutions that use as few weights as possible, otherwise, the solution found in the first task could consume all connections.

A weight pruning introduced by Han et al. (2015, 2017) shows that a model can have similar accuracy with fewer parameters if the proposed pruning-training process is incrementally executed. This process consists in nullifying some weights and retraining in the same data set until the accuracy cannot overcome a predefined threshold in the training. Those weights that were nullified are considered pruned and the neural network can solve the task only with the weights that are left. Therefore, Guo et al. (2018) used this process to select the minimal subset of parameters for each task. They also apply $L_1$ norm in the training phase to accelerate the pruning process. They showed that it is possible to resolve some tasks with less than 10% of total resources.

Masse et al. (2018) showed that with a random selection of 20% of the weight for each task (freezing the other 80%), the neural network can learn a sequence of 100 tasks without suffering much from CF in tasks like Permut-MNIST. For this approach to be successful, it needs to be associated with weights regularization like EWC or SI (Kirkpatrick et al., 2017; Zenke et al., 2017).

PackNet uses a binary mask of weights for each task (Mallya & Lazebnik, 2018). This causes an overhead of storage, as the number of tasks increases. To minimize the storage needed by masks, they proposed the use of an incremental mask. That is, if the model uses parameters $p_1, p_2$, and $p_3$ to learn the first task, it freezes these parameters and uses them together with others (not frozen) to learn a new task. To create the mask they also used the iterative pruning method of Guo et al. (2018). Serra` et al. (2018) extended PackNet by implementing an attention mechanism, called Hard Attention to the Task (HAT). It has a gate coupled to the output of the neurons that control the flow of information transmitted by the neuron. This mask is learned together with the task using a sigmoid function as a gate. As the sigmoid function produces a value between zero and one, they used a mechanism to force the result to be only zero or one, generating the mask. HAT was tested in eight sequential tasks, and it was able to learn them without CF of the initial one. Adel et al. (2020) equip each node of the model with three more variables to control the mask. The first one is binary and defines whether or not to be adapted, while the other two define the magnitude of the adaptation. These three variables are learned via variational inference.

Some tasks can share a mask, or at least a part of it, without causing the CF problem on past tasks while allowing the new task to be learned. It works on tasks that are similar. Ke et al. (2020) define that tasks $t_i$ and $t_{i+j}$, where $j$ represents a positive integer value, are considered similar if a positive knowledge transfer from $t_i$ to $t_{i+j}$ is observed. For this, a model trained from scratch in task $t_{i+j}$ is compared with a pretrained model in task $t_i$ fine-tuned to task $t_{i+j}$. If the fine-tuned model has a better performance they consider that a positive knowledge transfer has been observed. They also used HAT to find the masks.

Piggyback (Mallya et al., 2018) represents a departure from the above approach of selecting a subset of parameters for training specific to task $t_i$. The Piggyback method involves finding a mask that, when applied to a frozen pretrained model, such as a ResNet (He et al., 2016) that has been trained on the ImageNet (Russakovsky et al., 2015) data set, results in a solution for task $t_i$. The mask is used to decide which parameters of the model should be used for each task; however, the model weights are never changed. When the backbone is trained on a task in a different context than the tasks being evaluated, the results are not satisfactory as the new patterns cannot be learned. Zhai et al. (2020) extend this concept to GANs, creating a bank of filters that the model can use, instead of creating, in future tasks. Recently, Xue et al. (2022) show that these masks can also be used in models from Vision in Transformers (ViT) to avoid the CF.

Modular networks consist of multiple clusters, or modules, comprising densely interconnected neurons that have sparse connections with neurons in other clusters (Wagner et al., 2007; Lipson, 2007; Clune et al., 2013). In the context of RL, modular networks have been employed to address CF (Ellefsen et al., 2015). However, there is a lack of comparative results against established benchmarks, as the authors proposed a novel RL problem. They designed a distinct set of input neurons for each task, as depicted in Figure 11. The red neurons receive samples exclusively from one task, while the blue neurons receive samples exclusively from the other task. This arrangement enables the internal structure of the model to selectively activate different subsets of neurons for each task.

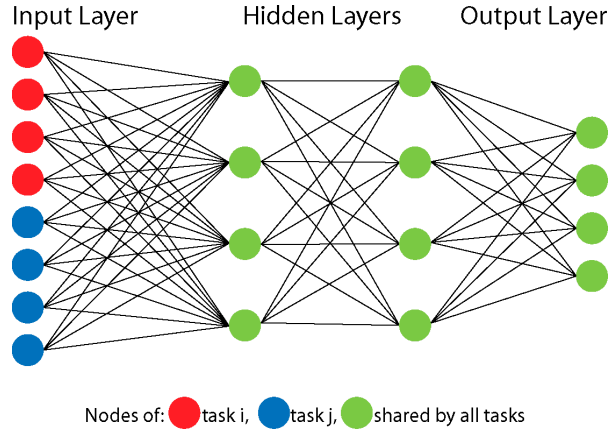

Figure 11: Figure adapted from Ellefsen et al. (2015). At the top, we can see the input layer. Each task has 4 values as input, thus, the first four neurons were used when training task $t_i$, and the other four were used on Task $t_{i+1}$.

A variation of modular networks, known as Diffusion-based Neuro-modulation, was introduced by Velez and Clune (2017) to identify task-specific modules within the model. The authors established connection points within the model to regulate modularity. For their evaluation, two additional input points were introduced, one for each task. The first point received a positive stimulus, while the stimulus at the other point was suppressed when processing samples from the first task. Conversely, the stimulus was inverted when processing samples from the second task. The authors solely tested their method on the specific problem they proposed, and its extension to other tasks is challenging due to the requirement for manual model design and configuration.

One potential solution to address the limited capacity of a model is to provide it with more capacity than necessary and enable every layer to contribute to the decision-making process for each task, thereby creating an ensemble model. An example of such an approach is the Incremental Adaptive Deep Model (IADM), which leverages the outputs from these layers as input to a shallow network. This shallow network employs an attention mechanism to make the final prediction (Yang et al., 2019).

Figure 12 provides an overview of the Hard Sub-networks category. Most studies within this category focus on exploring the optimal set of parameters to effectively solve multiple tasks. It is crucial to note that the ideal parameter set should also be minimal due to resource limitations. As the timeline progresses, all available resources (model parameters) become occupied, thereby rendering the model incapable of learning any additional tasks without experiencing CF. In an attempt to address this issue, some researchers have introduced a new category of strategies known as Dynamic Networks, which allow for model expansion when required. The dynamic networks category will be discussed in the upcoming section.

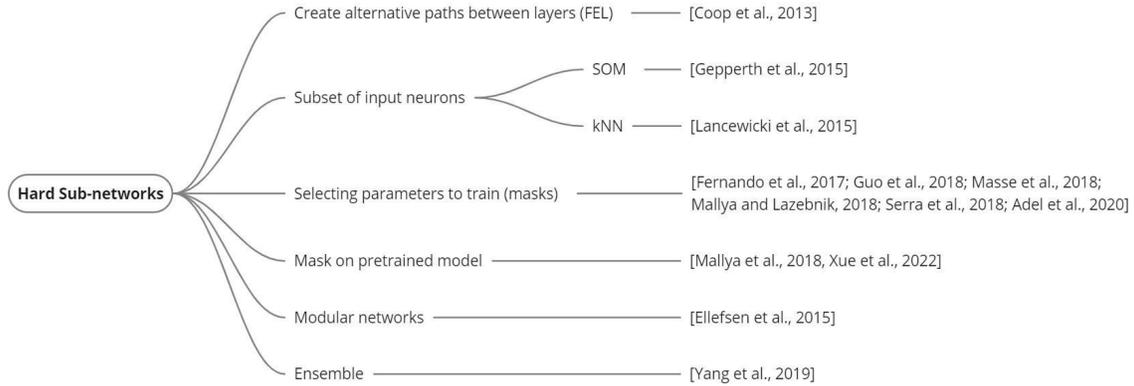

Figure 12: Overview of main approaches in Hard Sub-networks category. In Hard Sub-networks, each task will have permission to change just a unique subset of parameters. The works discussed in this section are clustered by their approaches.

### 3.4 Dynamic Networks

Dynamic Network category suggests that the CF problem is caused by the overlap of knowledge within neural network connections, but beyond this, it also suggests that a fixed number of resources (parameters) will not allow learning new tasks in perpetuity without forgetting previous ones. Therefore, the Dynamic Network techniques circumvent this problem, allowing the model to expand its capacity when necessary. The pioneering work in this idea was an analytical system based on the pseudoinverse of the input matrix, introduced by Serre (2002), which was used to identify the optimal parameters after adding a new neuron on the output layer. While this approach may effectively solve the problem of CF, it is limited to networks that follow the Extreme Learning Ma- chine (ELM) (Huang et al., 2006) architecture. ELMs are neural networks that consist of an input layer, a hidden layer, and an output layer. Because this approach relies on the specific structure of ELMs, it cannot be applied to DNN architectures that have more complex network structures.

In the context of DNN architectures, Rusu et al. (2016) start an investigation on this category by proposing Progressive Neural Networks (PNN). PNN replicates all the network's structure for each new task, initializing the weights of the new network with the most recently trained ones. In addition, each layer $l_k$ of task $t_i$ is connected to the input of layer $l_{k+1}$ in task $t_{i+1}$. PNN shows that expanding the capacity of the model is a possible way to handle CF in DNNs. The Expert Gate (Aljundi et al., 2017) method uses one model for each task; however, each model is equipped with an auto-encoder (AE) and is considered an expert to solve its task and has its parameters frozen after the training session. It is better than PNN to learn a new task because when a new model is required, the expert with better results on current training data is cloned. Therefore, it is not limited to the last model. Moreover, the AEs address the issue of identifying the task to which a specific sample belongs.

Another work that creates a new DNN model for each task is the Dual-Memory (Lee et al., 2017b). For each task, a new DNN is trained on data of the current task to produce an embedding. This embedding is concatenated with embeddings from all previous DNNs trained. This concatenation creates a rich representation of the data across all tasks learned so far, allowing the model to retain knowledge from previous tasks. Next, a shared shallow neural network selects a random subset of this concatenation to predict the class of a sample. The weights of a novel DNN are always initialized with the trained weights of the last DNN learned and frozen when the training session finishes. The main problem with these works is the inefficient use of resources, because they create a new entire DNN for each new task, even though the learning of the new task is faster as it takes advantage of the knowledge acquired by the old networks. Therefore, it required more research to minimize the increase of resources, looking for modular expansion, where growth is sublinear (Ostapenko et al., 2021).

Despite the inherent resource-intensive nature of Transformers architectures, certain researchers employ the Dynamic Networks approach to achieve enhanced accuracy and circumvent the challenge of CF in extended task sequences (Douillard et al., 2022; Hu et al., 2023). Leveraging the representational capabilities offered by Transformers, Douillard et al. (2022) propose to use a shared Transformer encoder (Vision in Transformer) to generate tokens from samples and increase the structure of the decoder to classify each task. More recently, expert modules to adapt the stimulus in internal layers of Transformers are created for each task and freeze them. In order to achieve a sub-linear growth rate, an effective strategy involves sharing modules across multiple tasks, as emphasized by Hu et al. (2023). It is crucial to acknowledge that the Transformer model under con- sideration is a pretrained model, whereby solely the modules undergo incremental learning during the process.

In order to mitigate the escalating rate of resource consumption, Ramesh and Chaudhari (2022) adopt a contrasting approach and suggest the utilization of a singular DNN for feature extraction, accompanied by multiple shallow networks, each dedicated to a specific task. By employing a shared feature extractor across all tasks, they propose the adoption of a pretrained model or the freezing of the feature extractor after training the initial task. During a training session, only the shallow network specialized in the current task undergoes updates. The inference is performed by aggregating the outputs of all classifiers, employing an averaging technique. The authors advocate that this approach proves effective due to its functioning as a Boosting classifier, ensuring improved classification performance while utilizing resources more efficiently.

Li and Hoiem (2018) propose to add only a new softmax layer in the model for each new task. This method is known as Learn without Forgetting (LwF). This method requires a specific training algorithm. Before starting a new training session, the model creates a set of soft labels for each data. A soft label is the probability distribution over multiple possible classes, in this case, it is generated by the softmax layers of the model. The soft label helps the model to generalize and is used to apply KD (Hinton et al., 2015) in the training session. When the layers generate the labels, they are known as Teachers. On the other hand, when the layers use these labels to acquire knowledge, they are known as Students. In the training session, it applies KD in those softmax layers while using the current task data to train the new softmax created.

The method described previously utilizes a shared feature extractor to generate an embedding representation prior to the softmax layer. However, one significant drawback of this approach is that the embedding can be altered as the network learns new tasks. As a result, it becomes difficult to maintain the original knowledge, since the embedding may undergo substantial modifications during training. This presents a challenge for developing effective algorithms to avoid CF following those methods (Rannen et al., 2017).

Other works focused on research metrics to indicate where and/or when the model has to increase. Cai et al. (2018) proposed a metric that measures how well a new task is fitted by filters in each layer of a CNN already trained. This metric is called Averaged Response Variance (ARV) and is calculated as ARV = $1/N \sum^{N} R_n$, where $N$ is the number of filters in each layer and $R$ is the variance of the filter outputs after processing the data set of the new task. They proposed to change the model's structure based on this metric. When the ARV value is large, they augment the number of filters to fit the new task knowledge. Ashfahani and Pratama (2019) show that other metrics can be used to make this decision. They proposed using overfitting and underfitting metrics. to modify the number of neurons in a layer and the number of layers if necessary. Although the accuracy results are comparable across different approaches, the growth of the model may vary depending on the task and its order. Thus, it is crucial to evaluate each scenario individually to determine which approach has the least growth rate.

Rather than relying on a metric, some researchers have opted for a trainable method where an auxiliary entity makes the decision. Neural Architecture Search (NAS) was used to expand a model until discovering the optimal structure to support the incoming task (Li et al., 2019). They consider three operations to find the optimal structure: i) add new filters to the layer, freezing the old ones;
ii) reuse the old frozen ones; and, iii) create a new layer from scratch. In parallel, (Xu & Zhu, 2018) trained an agent with RL, in an end-to-end fashion, to decide when and where to add nodes on a neural network to support new tasks. Those methods generally spend more time to reach a final version of the model, however, can find better structures (lower size). Hierarchical methods can also be used to decide when expanding the model. Tree-CNN (Roy et al., 2020) uses a tree structure to define its model. Each node can give a prediction or take it to another node to decide. Each node has its own CNN to predict. In addition, a child node is created when a new task is presented, and the node cannot distinguish the new task from any other that it has already learned. The main problem with this method is that when it makes a mistake at some intermediate level, the final inference will also be wrong.

A different research direction within this category involves the reuse of a "frozen" model and replication of only a specific portion of the original architecture. Unlike other approaches, this technique does not require any modifications to the overall structure of the DNN. Instead, the updated weights for only the relevant part of the model are retained, while the remaining weights remain un- changed. This approach can provide an efficient means of preserving the original knowledge in the network while allowing for the acquisition of new information for specific tasks. Imai and Nobuhara (2018) creates a new DNN with a structure identical to that of the pretrained one, however, with the weights initialized randomly. Thus, in a training session, the model can choose whether to use a $l_k$ layer from the pretrained DNN or from the recently created one. Therefore, only a part of the DNN needs to be stored and the technique can be applied to any architecture. The results show that the initial layers are generally taken from the pretrained DNN because it contains the base features. However, they do not present tests in a sequence of tasks with drift domain.

Based on the results of Imai and Nobuhara (2018), which show that the initial layers of a CNN learn low-level characteristics, Zacarias and Alexandre (2018b) proposed SeNA-CNN. This is a special case of CNN that has five convolutional layers. When the model needs to learn a new task, it just replicates layers three onwards for each new task and shares the first two. Different from PNN there are no connections between layers of different tasks. Both PNN and SeNA-CNN do not know what task a sample belongs to. To handle this issue in SeNA-CNN, Zacarias and Alexandre (2018a) published an improvement, where another CNN equipped with a *softmax* output layer decides which branch to use in prediction time.

An even more resource-efficient proposal was done by Xiong et al. (2018). They only store the statistics (mean and std) of the last layers, so in inference time, the weights are generated according to the task. To decide which task the sample belongs to, they use an auto-encoder for each task. While saving resources on the neural network, it creates new structures and information to store. It is important to point out here that when tasks share the first layers, the model presents favoritism in the first learned task. When the tasks have a similar domain it works fine, however, to a different domain it can deliver poor results.

Several studies have explored the utilization of methods and strategies within the Dynamic Net- works category to tackle CF in more challenging problem setups and contexts. For instance, Lee et al. (2020) work on Unbounded Task Incremental Learning proposing a framework that creates a new network, called Expert, to handle the unknown classes. As the data arrives in the stream, when the sample is not recognized by any Expert that already exists, the sample is stored in an ephemeral memory. When this memory is full, a new Expert is created and trained with this data, after this, the data is discarded. In the context of GANs, Verma et al. (2021) propose to integrate each parameter from two parts: i) shared one; and, ii) task-specific one. The former is fixed after the first task or gotten from a pretrained model. The latter is created for each new task, in such a way that, by adding the two parts, the model produces good results. To create distinguished task-specific parts, they apply the Kullback-leibler divergence constraints.

A typical challenge faced by Dynamic Network techniques is the requirement of knowing the task that a given sample belongs to. This problem also happens in some models of the Sub-network group too. Some works propose models equipped with their own task selector, but not all. Therefore, some authors focus on research methods to be used as oracles to decide which task each sample belongs to. Gepperth and Gondal (2018) propose to use a KNN as an oracle and just create a new output layer head to each task. This is resource-efficient; however, they only present tests on permutation MNIST data sets. Consequently, each task exhibits highly similar input statistics, i.e., there is no significant distribution shift across the tasks.

In summary, models within the Dynamic Networks category exhibit the ability to increase their size to accommodate new knowledge, with some models altering their structures while others create auxiliary structures. Figure 13 provides an overview of this category. The primary objective of these models is to learn new tasks while minimizing the increase in model size. However, some authors accept a polynomial increase in size to achieve more efficient learning of new tasks. Similar to the Sub-Networks category, distinguishing the task to which a sample belongs remains a challenge in Dynamic Networks approaches. This area of research is still actively evolving and holds promise for improving existing methods. However, a notable drawback of this category is the space required by the model, making it less suitable for scenarios such as edge computing or mobile ap- plications where space constraints are critical. In the subsequent section, we will discuss methods that combine strategies from multiple categories, aiming to leverage the strengths of each approach.

### 3.5 Hybrid Approaches

Researchers who often employ a hybrid approach combine two or more techniques described in previous sections to deal with CF effectively, thus leveraging the strengths of different strategies to achieve better performance. Certain combinations are relatively straightforward to implement, such as incorporating rehearsal methods with other techniques, and often yield enhancements in accuracy for previously learned tasks. The flexibility of hybrid approaches also allows them to

tailor their

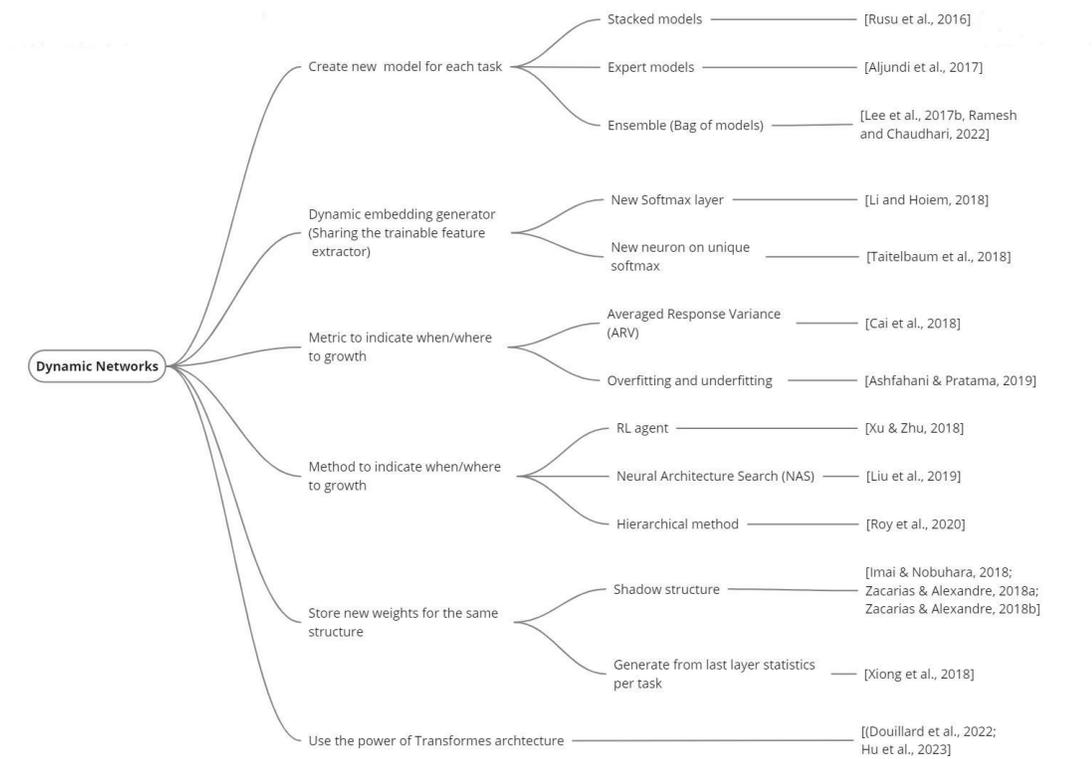

Figure 13: Overview of main approaches in Dynamic Networks category. In Dynamic Networks you can expand the resource (parameters) to arrange the new knowledge. The works discussed in this section are clustered by their approaches.

solutions to specific requirements and achieve a balance between mitigating forgetting and acquiring new knowledge.

### 3.5.1 Enhancing Distance-based using Rehearsal

Distance-based approaches encounter a significant challenge in effectively tracking class prototypes when the feature extractor undergoes weight modifications during training. These changes in weights result in shifts of the prototype vectors to different regions in the feature space, which can lead to confusion in the final classifier's decision-making process. To address this issue, rehearsal techniques can be employed by storing a subset of samples and their corresponding feature vectors. By incorporating rehearsal, the feature vectors of these stored samples can be closely monitored during the learning of new tasks, enabling the detection of any distance shifts caused by weight modifications. ICaRL pioneered the combination of rehearsal and distance-based methods to tackle CF, effectively maintaining accurate class representations despite weight modifications and proto- type shifts (Rebuffi et al., 2017).

Rehearsal, as demonstrated by Liu et al. (2020), can be effectively combined with a Distance-based and Knowledge Distillation (KD), in a meta-learning context, to enable adaptation to new tasks while minimizing the interference with previously learned tasks. This integration is complemented by the use of a coreset, which helps maintain consistent embeddings of stored samples. The pro- posed model, presented by Liu et al. (2020), consists of two main components: an extractor and a classifier. At the start of the training session, a temporary copy of the extractor is created with its parameters frozen. Throughout the training, both the frozen extractor and the active one produce identical embeddings for the same input data. This alignment is crucial to ensure that the classi- fier achieves high accuracy when predicting new classes during the fine-tuning process. The entire process takes place within the meta-learning inner loop.

Furthermore, in an Online Incremental Learning setup where the model lacks task ID knowledge and must learn in a single pass, a Distance-based approach augmented with a Mini-rehearsal technique can be employed to mitigate CF. To address the challenge of limited samples per class, the MixUp technique introduced by (Liang et al., 2018) can be utilized for data augmentation. By combining these strategies, including Distance-based, KD, Mini-rehearsal, and MixUp, a model can adapt to new tasks, retain past knowledge, and mitigate CF while overcoming limitations associated with this setup.

Distance-based approaches benefit from using cosine distance classifiers, which have shown greater resilience to CF compared to softmax classifiers (Simon et al., 2021; Davari et al., 2022). However, this advantage holds true only when the embeddings belonging to the same class can be effectively clustered in the latent space and remain sufficiently distant from other class clusters. To address this requirement, Mai et al. (2021b) propose a solution that involves storing a coreset of samples and applying Contrastive loss during training. By leveraging the coreset and Contrastive loss, the aim is to encourage the embeddings to automatically cluster in a way that facilitates better separation between classes. Thus, the model can better organize the embeddings in the latent space, resulting in improved class separation and enhanced resilience to CF.

In a highly challenging setup, such as Online Incremental Learning combined with an Unbounded Task scenario, the utilization of rehearsal techniques becomes crucial for preserving previous knowledge in Distance-based methods. To address this, De Lange and Tuytelaars (2021) pro- pose a memory-based approach that consists of two components: samples and prototypes. Memory storage effectively handles the unbalanced data stream that arises in such setups. The assumption is that prototypes will change over time as new information is received. Therefore, a novel loss function is introduced to guide the feature extractor in generating embeddings that are closer to the class prototype while being farther away from other prototypes. This encourages better separation between classes and improves the model's ability to retain previous knowledge. As the samples in the memory are replaced with new ones, the prototypes evolve accordingly, adapting to the changing data distribution. This dynamic updating of prototypes ensures that the model remains up-to-date and capable of handling evolving tasks and unbounded learning scenarios effectively.

The analysis of geometric relations among classes can be examined from multiple perspectives through the utilization of Mixed-curvature Spaces, as proposed by Gao et al. (2023). Their approach involves creating distinct spaces for each task and projecting samples onto all relevant spaces during inference. By leveraging a coreset, their objective is to ensure that new projections do not introduce confusion with classification from previous spaces. Similarly, Ma et al. (2023) employ a projection technique for sample embeddings. However, they adopt a Voronoi Diagram as the designated space. Specifically, when a new class emerges, they partition a cell within the diagram. One portion of the split is dedicated to retaining existing knowledge, while the other part accommodates the new class. This process, facilitated by the utilization of a coreset, entails strategic decisions on cell selection for partitioning and ensuring the preservation of a structured framework that mitigates the challenges posed by CF.

In scenarios where the total number of classes is known in advance, yet the classes must be learned sequentially, it is feasible to predefine a distinct region for each class and solely train a projection layer, as proposed by Yang et al. (2023). It is important to note that the underlying embedding generation model remains fixed throughout this process, and it can either be a pretrained model or be frozen after the initial task. Since the projection layer necessitates updates with each new task, the inclusion of a coreset becomes imperative to retain the knowledge on how to appropriately project each class into its respective space, thereby effectively addressing the challenge of CF.

### 3.5.2 Enhancing Distance-based using Dynamic Networks

Hybrid approaches that combine Dynamic Networks with Distance-based methods offer another effective solution. The best example is the Encoder-Based Lifelong Learning approach, which creates a new classification layer for each task and tracks changes in class embeddings (Rannen et al., 2017). In this method, an auto-encoder is trained for each task, allowing the preservation of embeddings and enabling the fixation of the classification layer for old tasks through knowledge distillation.

The use of auto-encoders in this approach serves two key purposes. Firstly, it allows the embed- dings of one class to be stationary over the model lifespan. Secondly, the auto-encoders can serve as oracles, providing guidance to the classifier in the model's lifelong learning process. This selection is based on the distances between the preserved embeddings, ensuring accurate and task-specific classification.

By combining Dynamic Networks with Distance-based methods and leveraging the Encoder-Based Lifelong Learning approach, a model can adapt to new tasks while retaining knowledge from previous tasks. This hybrid approach offers a comprehensive solution that effectively handles lifelong learning scenarios mitigating CF. However, since the approach relies on training a separate auto-encoder for each task, the memory footprint of the model grows as new tasks are introduced. Therefore, the auto-encoder structure has to be much lower than the original model to be a viable approach.

### 3.5.3 Enhancing Dynamic Networks using Rehearsal

One significant limitation of Dynamic Networks methods lies in their elevated growth rate. To mitigate and control this growth, a prudent approach involves maintaining a coreset, which enables the model to modify certain parameters while preserving knowledge of previous classes. Consequently, the model can effectively acquire new knowledge with reduced resource requirements, such as fewer neurons and layers.

Lüders et al. (2017) propose to use an Evolvable Neural Turing Machine (Greve et al., 2016) (ENTM). In this model, a tape is added to store contextual information. This tape has two heads, one to read and the other to write, which are controlled by a neural network. They used an approach called Neuroevolution of Augmenting Topologies (NEAT) (Stanley & Miikkulainen, 2002) to produce this neural network. NEAT uses evolutionary algorithms to increase the size of the neural network, so it starts with a shallow network and evolves it to a more robust one until it finds an acceptable solution. In addition, ENTM stores the contextual information that allows the neural

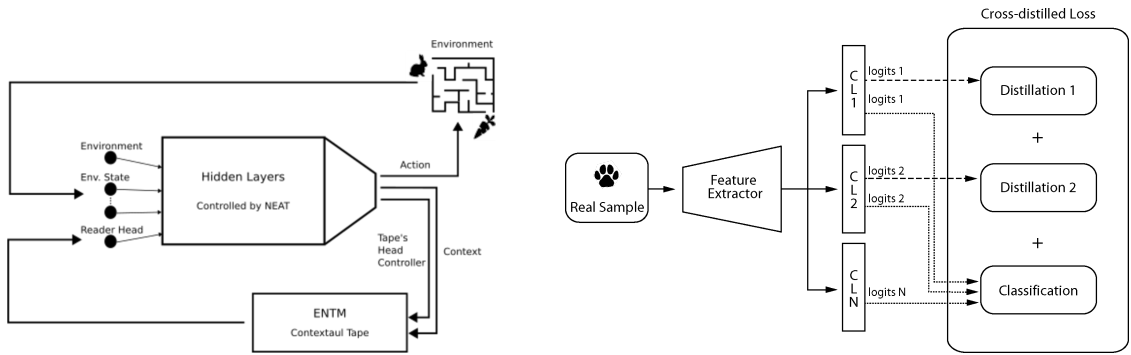

(a) Architecture proposed by Lüders et al. (2017).  (b) Castro et al.'s (2018) model training.

Figure 14: On the left side (a), a neural network with fixed input and output layer controls the agent actions taking as input the last reward, the current environment state, and the last contextual information maintained in ENTM's tape. In addition, the neural network produces a new context to be stored in ENTM's tape and controls its heads. The amount and the shape of the hidden layers are controlled by a NEAT algorithm. On the right side (b), the classification layers from the old tasks produce logits that are used for distillation and classification while the classification layer from the current task produces logits that are involved only in classification.

network to remember things. The left side of Figure 14 illustrates the proposed architecture. This model is hybrid, ENTM is its complementary learning system, and the NEAT module makes its neural network grow dynamically. Unfortunately, they only tested this proposal in an RL context and required a handcraft-designed input and output by task. The input needs to receive three pieces of information: i) the new state; ii) the reward (environment output); and, iii) the contextual information in the reader's head tape. The network's output produces: i) the action (environment input);

ii) a context representation to be written in the tape; and, iii) controllers to the tape's heads. Besides that, they do not compare their results with other works, just demonstrating that this model can solve the RL problem defined by Ellefsen et al. (2015).

Castro et al. (2018) expand the model adding a new classifier to each task and using an approach based on herding selection to maintain samples on the coreset. In contrast to ICaRL (presented in Section 3.5.1), they train a parametric classifier together with the feature extractor. The right side of Figure 14 presents the process of training. For each new task, a new head classifier is added to the model and trained with cross-entropy loss, while the oldest ones are trained with distillation loss using KD. They store a part of the older task data but also grow the network structure by adding a new head to each task. To handle the problem of the imbalanced data set, they apply a post-training fine tune with samples from memory and a subset of samples from the current task such that all classes have the same number of samples.

Ostapenko et al. (2019) extend DGR (Shin et al., 2017) adding the capability of expanding the generator model, creating a hybrid model from Pseudo-rehearsal and Dynamic Networks. After a training session, the generator parameters are frozen, then in the next session, more resources (layers and neurons) are added. Due to this, it requires storing a mask to identify which parameters belong to each task. Instead of generating pseudo-samples, Joseph and Balasubramanian (2020) propose to use a Variational Autoencoder (VAE) to generate weights to create an ensemble of models, all of them with the same structure. These weights are considered a kind of rehearsal of old knowledge. Although the method proposed by Hu et al. (2019) also has a weight generator module, the technique proposed by Joseph and Balasubramanian (2020) differs from them because they create an ensemble of models.

RL agent was used as a neural structure search method to expand the model in a Task Incremental Learning setup (Qin et al., 2021). It considers as the environment the current task data, a continual learner, a knowledge repository, and the coreset. The knowledge repository contains the parameter's value of old tasks. Thus, the agent uses the environment to create a new learner, minimizing the forgetting ratio (validating it with the coreset) and maximizing the accuracy of the current task.

An additional prevalent challenge encountered in Dynamic Networks methods pertains to the determination of the task to which a sample belongs during inference. The application of a coreset can serve as a valuable aid in alleviating this predicament as well. Dynamically Expandable Representation Yan et al. (2021) expands the model creating a new feature extractor for each task and having a unique classifier. Once the extractor is learned, it is frozen. The classifier is always modifying its parameters to learn new tasks, so it needs to store a coreset to avoid the CF. The classifier uses as input the concatenation of features produced by all extractors. Therefore, the representation size of each sample increases when a new task is learned.

An alternative approach to address the challenge of identifying the task associated with a sample in Dynamic Networks methods is to employ multiple classifier heads, each dedicated to a specific task. By assessing the confidence levels of each classifier, the one exhibiting the highest confidence can be selected. Kim et al. (2022) propose to use the coreset, containing data from previous tasks, as a negative class within the current head classifier. Consequently, each classifier can function as an effective out-of-distribution detector, enabling the appropriate head selection during inference.

Both Dynamic Networks and Rehearsal techniques suffer from the disadvantage of increased storage requirements compared to the original model. Dynamic Networks necessitate storing additional parameters, while Rehearsal involves storing samples from a coreset. The increased space de- mands should be carefully considered in terms of resource utilization and memory constraints. Zhou et al. (2023) introduces a baseline framework for this hybrid category, which takes into account the overall memory consumption, additional parameters, and the presence of samples within the core- set. This framework is designed to strike a delicate balance between the number of retained samples in the coreset and the expansion of the model itself. Through their empirical analysis, the authors note that expanding the model at the deeper layers yields substantial advantages in terms of both accuracy and space requirements. Notably, this proposed baseline outperforms, in accuracy and space requirements, numerous techniques discussed in this section.

3.5.4 ENHANCING SUB-NETWORKS USING REHEARSAL

The application of Incremental Learning using Sub-networks techniques holds particular significance in scenarios characterized by limited computing resources, such as edge computing. Notably, the integration of a coreset, which retains select samples from previous tasks, can play a crucial role in facilitating the discovery of improved and more shareable sub-structures between tasks (Wang et al., 2022e). Recent studies have highlighted that it may not be necessary to regularize all model weights to mitigate CF, as focusing on regularization solely in the last layer can suffice. This approach, known as functional regularization, has been effectively employed by Titsias et al. (2020) and Pan et al. (2020), utilizing a Gaussian Processes (GP).

One drawback of functional regularization using GPs is its reliance on storing a large number of samples from the most recent tasks. Consequently, a hybrid approach combining regularization (identifying sub-networks within the last layer) with mini-rehearsal techniques should be favorable. It is important to note that the choice of mini-rehearsal technique varies across different studies. The results achieved through this approach exhibit similarities to methods that adopt regularization across all model weights while being computationally more efficient.

Another noteworthy technique is Neural Calibration for Online Continual Learning (NCCL) proposed by Yin et al. (2021). NCCL extends the concept of Elastic Weight Consolidation (EWC) by introducing two mechanisms to modulate the input and output signals of DNNs, referred to as neuron calibration. The parameters of this module are learned to strike a balance in the stability-plasticity dilemma, aiming to discover sub-modules within the model that can effectively solve each task. However, it still requires storing a portion of the training data from each learned class to aid in the learning process.

Sarfraz et al. (2023) employ two key techniques to encourage the formation of Sub-networks using Rehearsal techniques: Sparse Activations and Semantic Dropout. Sparse Activations utilize the k-Winner-Take-All (k-WTA) activation function, enabling the selective propagation of the top $k$ units with the highest activations to subsequent layers. On the other hand, Semantic Dropout promotes dropout in different neurons when learning new tasks, leveraging the activation patterns of neurons. This approach proves particularly advantageous when dealing with uncorrelated tasks.

Incorporating Sub-networks and Rehearsal techniques offer promising avenues for continual learning in environments of fixed space resources. They provide mechanisms to identify and lever- age task-specific sub-structures within the model, thereby facilitating better retention of previously acquired knowledge while adapting to new tasks. Nevertheless, it is crucial to address the challenge of storing training data to ensure the effectiveness of these techniques.

### 3.5.5 GROWTH SUB-NETWORKS USING DYNAMIC NETWORKS

The combination of Dynamic Networks and Sub-networks offers a solution to the limited resource challenge encountered by Sub-networks methods. One method in this regard is the Dynamically Expandable Network (DEN) introduced by Yoon et al. (2018). DEN aims to determine the minimum set of connections required by a Neural Network to learn a new task. If the existing connections are insufficient, DEN dynamically expands the model's resources.

DEN operates through three distinct phases: training, expanding, and avoiding semantic drift. In the training phase, the model undergoes initial training using $L_1$ regularization, which promotes weight sparseness. When a new task is introduced, an output node corresponding to the task is added. Subsequently, a pre-training step takes place, where only the last layer remains unfrozen, adjusting only the weights of the newly added node. Since the remaining parts of the model also maintain sparse connections, a Breadth-First Search (BFS) algorithm is employed to establish a path $S$ that connects every output node to the model inputs. Afterward, all neurons except those belonging to path $S$ are frozen, and the model undergoes conventional training, minimizing the loss associated with the specific task.

During the expanding phase, the model checks if the loss for the new task has reached a pre-defined acceptable level. If not, $k$ neurons are added to all layers and another round of training takes place using $L_1$ regularization to promote sparsity and discard unnecessary neurons.

Finally, in the avoiding semantic drift phase, the model examines if any neuron in the original path $S$ has undergone significant changes. If so, it indicates that the neuron is prone to semantic drift with respect to some previous task. Then, the neuron is duplicated: one of them preserves the old value and is removed from path $S$, while the other remains for the current task.

By dynamically adjusting the model's architecture and selectively expanding and modifying connections, DEN addresses the resource limitations faced by Sub-networks methods. This adaptive approach allows the model to accommodate new tasks while mitigating interference with prior knowledge and ensuring efficient resource utilization.

Another hybrid approach, proposed by Hung et al. (2019), also addresses the challenge of limited resources by identifying sub-structures for each task and expanding them when necessary. Their method involves an iterative process consisting of three phases. In the first phase, the model learns the task by utilizing unfrozen neurons while applying a mask to the frozen ones. The second phase involves iterative pruning of the non-frozen neurons to maintain the model's capacity. This pruning process helps optimize the resource allocation and ensures that only the most relevant connections are retained. Lastly, if the model fails to achieve the predefined minimum accuracy, it undergoes expansion and the training process recommences for the specific task. This expansion allows the model to acquire additional resources and adapt to the task's requirements. Once the pre-defined accuracy threshold is reached, the mask utilized during the learning phase is saved, and the corresponding weights are frozen.

By iteratively learning, pruning, and expanding the model's sub-structures, this method provides a hybrid solution that effectively manages resource limitations. This approach allows the model to dynamically allocate resources based on task-specific requirements while preserving the knowledge gained from previous tasks.

This combination takes the advantage of Sub-networks category of encapsulating the maximum number of tasks in the same network structure while avoiding the resource limitation. On the other hand, it only grows when it is really necessary, handling the problem in the Dynamic Networks category of increasing the size in a linear way. Therefore, it is considered an excellent choice, however, there are several environments and setups that it does not work, for instance, in Unbound Task Incremental setup or edge computing environment. In the general case, the accuracy of techniques in this category is worse than the techniques from the Rehearsal category.

### 3.5.6 Combining Sub-networks, Dynamic Networks, and Rehearsal

The Else-Net model proposed by Li et al. (2021) combines features of Sub-networks, Dynamic Networks, and Rehearsal. In this model, each layer consists of multiple knowledge blocks that process the input. A gating mechanism selects the most important block, generating an embedding for the next layer. When a new task needs to be learned, a temporary block is created. If the gate identifies it as important, it is retained for future tasks; otherwise, it is discarded, and other modules have their parameters adjusted. Additionally, a small portion (10 percent) of samples from each learned class is stored in the coreset to mitigate forgetting.

A notable difference from other methods discussed earlier is that the presence of a coreset allows for weight adaptation in a portion of the model without causing significant forgetting. This flexibility makes the Else-Net model applicable to setups beyond Task Incremental Learning, where the task identifier is known during both training and evaluation phases.

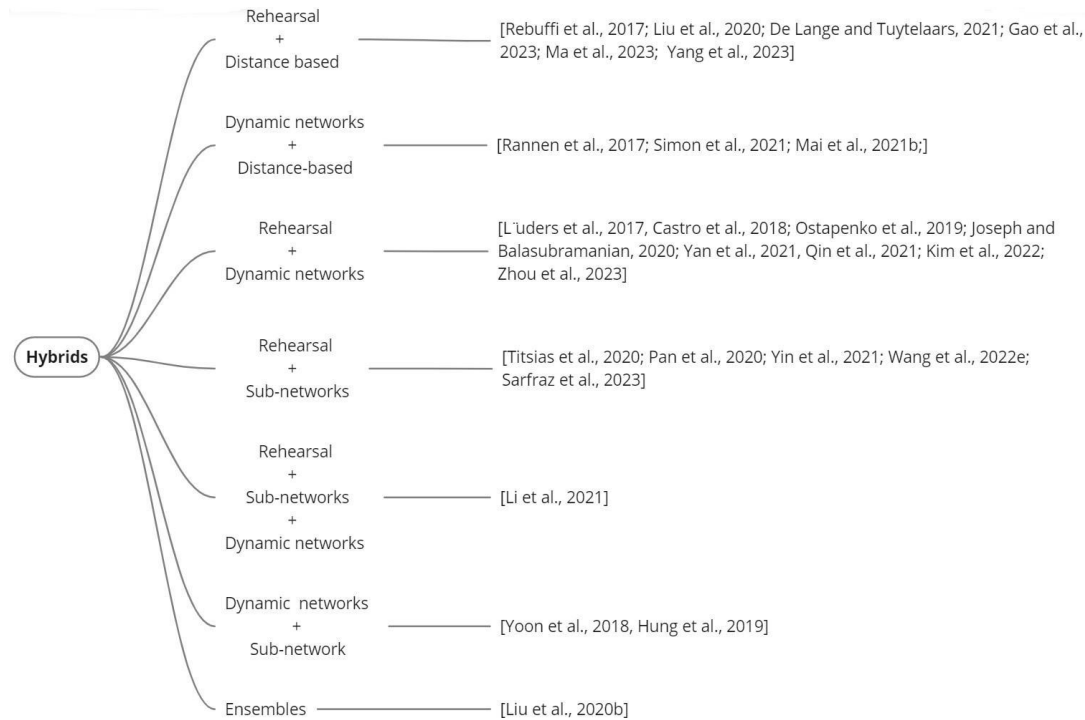

Figure 15: Overview of main methods in the Hybrid approaches category. The works discussed in this section are clustered by similarity.

Figure 15 provides an overview of approaches that combine multiple categories to address CF. Each category has its own strengths and limitations. It is evident that no model can learn an infinite number of tasks with a fixed parameter count. Conversely, creating a separate model for each new task is impractical. Storing all previously trained samples is also infeasible due to storage space and retraining time constraints, but preserving key samples appears to enhance the retention capacity. Comparing these approaches is challenging due to the lack of consensus regarding metrics, datasets, and testing methodologies.

## 4. Final Considerations

Catastrophic forgetting, initially identified as a problem in machine learning by McCloskey and Cohen (1989), refers to the phenomenon where a model loses previously acquired knowledge when learning new tasks. French (1999) further investigated this issue and found that it is particularly pronounced in connectionist models like Neural Networks. Nowadays, despite the significant ad- vancements in DL, addressing the challenge of CF remains an ongoing problem in the field, as evident from our comprehensive review. While pretrained networks in Computer Vision, and more recently in Natural Language Processing, exhibit reduced susceptibility to CF, relying solely on pretraining can lead to poor generalization for new tasks that differ significantly from the initial task. To overcome this limitation, it is crucial to ensure the diversity and heterogeneity of the dataset used for the initial task.

In our review, we explored various methods and techniques to mitigate the effects of Catastrophic Forgetting in DL. To facilitate the comprehension and analysis of these approaches, we introduced a taxonomy that classifies them into four main categories. This classification framework enhances our understanding of the different strategies proposed in the literature, enabling future researchers to categorize and compare their effectiveness in addressing CF.

The first category we reviewed is Rehearsal, which was originally introduced by Robins (1993). Rehearsal involves reprocessing previously encountered data when learning a new class. However, in practice, this approach is often infeasible due to the accumulation of a large volume of data over time. Within this category, we further divided it into two groups: pseudo-rehearsal and mini- rehearsal. Pseudo-rehearsal techniques aim to overcome the storage issue by generating simulated data on demand instead of storing the actual samples. This approach allows for more efficient use of resources while still providing the model with relevant information from previous tasks. On the other hand, mini-rehearsal approaches address the problem by storing a limited amount of samples from each class. These stored samples are carefully selected to ensure that the model does not suffer from catastrophic forgetting, even with this restricted subset of data.

During our review, we observed that there is room for improvement in sample selection algorithms for generating simulated samples on demand. By enhancing the accuracy and informative- ness of these simulated samples, we can potentially enhance the effectiveness of pseudo-rehearsal techniques. Additionally, we encourage the utilization of Generative Adversarial Networks (GANs) for sample generation, as they have shown promise in producing more informative and diverse samples. Overall, the Rehearsal category presents interesting avenues for addressing catastrophic forgetting, and further advancements in sample selection algorithms and the use of GANs can con- tribute to more effective and efficient rehearsal-based techniques.

The second category we reviewed is distance-based approaches. This category focuses on non-parametric classifiers, such as cosine classifiers, which utilize class representations generated by deep models, as demonstrated by Rebuffi et al. (2017). This approach helps address the issue of bias in the latest classes by leveraging the class representations. However, it necessitates the use of an effective embedding generator that minimizes overlap among representations of different classes while accurately approximating the representations of the same class. Notably, unsupervised learning and ranking losses have emerged as the most effective strategies within this category, as evidenced by recent research.

The primary challenge encountered in the distance-based category is maintaining the consistency of the class representations when new training sessions occur. As new tasks or data are introduced, it is crucial to ensure that the representations remain reliable and informative. Over- coming this challenge is essential for preventing CF and enabling effective classification on both old and new tasks. To enhance the distance-based approaches, future research could focus on devel- oping more robust and accurate embedding generators that can better capture the subtle differences between classes. Additionally, investigating novel unsupervised learning techniques and refining ranking loss functions can further improve the effectiveness of this category. By addressing these challenges, we can enhance the performance and stability of distance-based methods for mitigating catastrophic forgetting.

The third category we reviewed was sub-networks, which addresses the issue of CF by minimizing changes in neuron connections during the learning of new tasks. In Sub-network the CF problem arises when the value of a neuron connection, which has already learned a task $t_i$, changes in favor of learning a new task $t_{i+1}$. Thus, the main objective of sub-networks approaches is to mitigate such changes. Notably, Elastic Weight Consolidation (EWC) by Kirkpatrick et al. (2017) and PathNet by Fernando et al. (2017) are pioneering works in this category, although they may not achieve the best performance results.

We can classify similar approaches to EWC as Soft Sub-networks since they allow changes in all parameters of the neural network while penalizing changes in connections that are important for previous tasks. On the other hand, works similar to PathNet fall into the category of Hard Sub-networks, where the values of neuron connections, specifically those crucial for learning certain tasks, are frozen after training to prevent catastrophic forgetting. However, it is important to consider the limitations of Hard Sub-networks, such as their restricted number of parameters, which restricts the total number of tasks that can be learned. Additionally, the usage of an oracle is necessary for PathNet to determine the parameters to be used for each sample during inference. In contrast, Soft Sub-networks do not face this limitation, but they show lower performance in scenarios where domain tasks drift. Therefore, it is crucial to explore improved techniques for selecting parameters as an oracle in such contexts, as it holds the potential for significant advancements in the sub- networks category.

The final category we reviewed was dynamic networks, which aims to address the limitations of limited resources by dynamically increasing the size of the neural network as required. The representative work in this category is Progressive Neural Networks (PNN) by Rusu et al. (2016), which overcome constraints by allocating more resources for each new task. For instance, PNN demonstrates the creation of a completely new neural network for each task, resulting in improved performance.

The key research contribution in this category is to determine when and how to expand the network structure effectively. Finding optimal strategies for increasing the network's capacity is crucial. Additionally, exploring approaches to reduce the network structure could also lead to significant improvements. This is because as we gain knowledge in solving tasks and subsequently learn new incremental tasks, the previously learned tasks tend to become easier. However, a common challenge encountered here is the selection of parameters for inference. Determining the appropri- ate parameters set for a given sample during inference is critical for achieving optimal performance. Hence, it is recommended to investigate techniques that can serve as an oracle for parameter se- lection. Advancements in developing effective oracle techniques would benefit both the dynamic networks and sub-networks categories.

Lastly, hybrid models offer a promising approach to mitigating CF. Rehearsal techniques can be effectively combined with methods from other categories to address this challenge. However, there is a lack of comparative studies on the performance of these combined approaches. The combination of sub-network and dynamic network methods may not seem intuitive at first, as the former focuses on identifying task-specific sub-structures within the model, while the latter expands the model's capacity to accommodate new knowledge. Nevertheless, integrating elements from different categories can help reduce resource limitations encountered by sub-networks. In a study by (Liu, Parisot, Slabaugh, Jia, Leonardis, & Tuytelaars, 2020b), a mixture of techniques in the form of ensembles, such as EWC, IMM, and ICaRL, was explored. This approach aims to leverage the strengths of each model to enhance resistance against CF. However, it is worth noting that ensembles can come with a high computational cost.

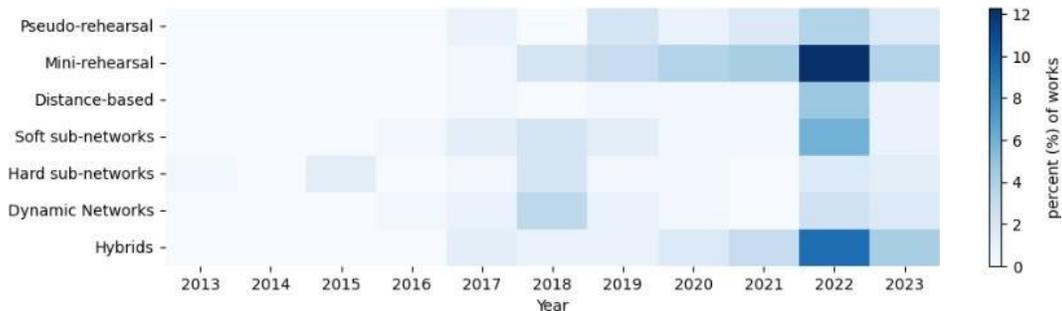

Figure 16: Heatmap showing the distribution of works sorted by categories. It depicted the distribution of all surveyed works in all years. We can note that in the last three years, the field gained a lot of attention accumulating most of the work. Moreover, the Rehearsal and Hybrids categories have a major concentration of works.

To date, our survey has examined a comprehensive selection of over 230 studies, primarily focusing on works published after 2013 when Deep Learning gained significant attention. We aimed to capture the most recent advancements in the field. Figure 16 presents the distribution of works across the proposed categories over the years. Notably, during the initial period from 2013 to 2015, the Sub-networks category dominated the landscape as the primary focus for mitigating CF. How- ever, in subsequent years, we observed a shift in research interest, with increasing contributions from other categories, resulting in a more balanced distribution. This diverse distribution implies that there is no clear-cut path or prevailing approach to achieving continuous learning while effectively addressing the challenge of CF.

We have observed that in the last two years, a majority of the works have adopted some variant of the Mini-rehearsal technique. It is worth noting that several studies have utilized techniques similar to a coreset to address CF, such as Incremental Semantic Segmentation (Oh et al., 2022; Yu et al., 2022). However, during our review, we discovered that not all papers proposed novel techniques specifically targeting CF. Instead, many of them applied existing techniques to various applications, making it impractical to cite each individual work. To overcome this limitation, we recommend referring to a curated repository[1] that provides an extensive and regularly updated list of papers on Incremental L0.49earning, where the majority of them attempt to address CF. In this review, our focus was on works that have made significant contributions to each evolving category.

More recently, in the domains of Natural Language Processing and Computer Vision, the utilization of prompts combined with few-shot learning in Transformers (ViT) has shown promise in addressing incremental learning without retraining the model, thereby mitigating the risk of forget- ting (Wang et al., 2022b, 2022d; Smith et al., 2023; Hu et al., 2023; Villa et al., 2023). However, it is important to acknowledge that this approach is not a definitive solution to CF, as its effectiveness is confined to specific domains. Furthermore, the current computational demands of most transformer models make them unsuitable for regular updates of their parameters to solve new tasks on a regular basis. Furthermore, it should be noted that most of these proposals present good results only in the context of Task Incremental Learning, making their practical implementation unfeasible in more realistic scenarios, such as Online or Unbound Incremental Learning.

---

1. `https://github.com/xialeiliu/Awesome-Incremental-Learning`

Although catastrophic forgetting (CF) is a well-established problem, there remains considerable divergence in testing protocols. Therefore, it is crucial for the field to adopt a standardized test plat- form that enables fair comparisons among new models and existing ones (Lomonaco et al., 2021). As a final consideration, it is important to recognize that the choice of the best model, technique, or framework will depend on the specific context and problem being addressed. Furthermore, it is likely that solving new complex problems will require the integration of multiple categories rather than relying solely on a single approach.

## Acknowledgments

The present work is the result of the Research and Development (R&D) project 001/2020, signed with Federal University of Amazonas and FAEPI, Brazil, which has funding from Samsung, using resources from the Informatics Law for the Western Amazon (Federal Law nº 8.387/1991), and its disclosure is in accordance with article 39 of Decree No. 10.521/2020. And, is a result of the R&D project (Tracking Platform), carried out by Sidia Instituto de Ciêncincia e Tecnologia, in partnership with Samsung Eletrônica da Amazônia Ltda.

ALEIXO, COLONNA, CRISTO & FERNANDES

M.,